\documentclass{article}

\usepackage{arxiv}
\usepackage[utf8]{inputenc} % allow utf-8 input
\usepackage{textcomp}
\usepackage[T1]{fontenc}    % use 8-bit T1 fonts
\usepackage{hyperref}       % hyperlinks
\usepackage{url}            % simple URL typesetting
\usepackage[super,sort&compress,comma]{natbib} 
\usepackage{booktabs}       % professional-quality tables
\usepackage{amsfonts}       % blackboard math symbols
\usepackage{nicefrac}       % compact symbols for 1/2, etc.
\usepackage{microtype}      % microtypography
\usepackage{lipsum}
\usepackage{graphicx}
\usepackage{doi}
\usepackage{amsmath}
\usepackage{amssymb}
\usepackage{pifont}% http://ctan.org/pkg/pifont
\newcommand{\cmark}{\ding{51}}%
\newcommand{\xmark}{\ding{53}}%
\usepackage{multirow}
\usepackage{placeins}
\usepackage{chemformula}
 
\newcommand{\vx}{\mathbf{x}}

\newcommand{\R}{\mathbb{R}}

\newcommand{\Loss}{\mathcal{L}}

\title{
    \textbf{Data Fusion and Contrastive Alignment for Unconstrained IR Molecular Structure Elucidation}
}

\author{
    Ethan J. Mick \\
    Electrical Engineering and Computer Science\\
    University of Missouri\\
    Columbia, MO 65211 \\
    \texttt{ejmcmk@missouri.edu} \\
    \And
    Campbell A. Sweet \\
    Chemical and Biomedical Engineering\\
    University of Missouri\\
    Columbia, MO 65211 \\
    \texttt{campbellasweet@missouri.edu} \\
    \And
    Matthias J. Young \\
    Chemical and Biomedical Engineering\\
    University of Missouri\\
    Columbia, MO 65211 \\
    \texttt{matthias.young@missouri.edu} \\
    \And
    Derek T. Anderson \\
    Electrical Engineering and Computer Science\\
    University of Missouri\\
    Columbia, MO 65211 \\
    \texttt{andersondt@missouri.edu} \\
}
 
\date{}   % suppress date for preprint; arXiv adds its own

\renewcommand{\shorttitle}{\textit{arXiv} Template}
 
\begin{document}
\maketitle
 
% ==============================================================
\begin{abstract}

%\vspace{-0.75em}

Automated molecular structure elucidation from infrared (IR) spectroscopy data has seen significant advancements in recent years, but its broad applicability is limited by a reliance on pre-determined chemical formulas provided as auxiliary model inputs. This limits model predictions to isomer identification rather than full molecular structure prediction. Although transformer models have been shown to identify molecular isomers with high accuracy, their reliability for unconstrained structure elucidation is comparatively low and poorly understood. In this work, we propose and evaluate key modifications to the traditional encoder-decoder transformer. To better address the vast chemical space of the unconstrained problem, we implement a novel Mixture-of-Experts (MoE) decoder module that utilizes non-additive aggregation via linear-order statistics and the Choquet integral. We further modify the transformer to utilize these non-additive operators when aggregating spectral representations as well. Together with an auxiliary contrastive alignment loss term, these enhancements improve Top-K prediction accuracy by over 10 percentage points compared to baseline IR-only models. Through sub-structure fragment analysis of molecular predictions, we further confirm that infrared spectra encode the vast majority of relevant chemical information, implying that the higher performance of isomer-ranking models is largely due to underrepresented or overlapping absorption bands for molecules in the explored chemical space. Ultimately, by demonstrating the efficacy of automated molecular structure elucidation from measured IR spectra, this work serves to significantly broaden the utility of AI in analytical chemistry.

\end{abstract}
 
% ==============================================================
\section{Introduction}
\label{sec:intro}

The determination of molecular structure from spectroscopic measurement is among the foundational problems of analytical chemistry. In practice, full structure elucidation is rarely accomplished by a single technique; it proceeds instead as a sequential, multi-instrument workflow in which orthogonal analytical methods contribute complementary layers of structural evidence. High-resolution mass spectrometry (MS) establishes molecular formula and an exact mass; nuclear magnetic resonance (NMR) spectroscopy resolves atomic connectivity, stereochemistry, and chemical environment; and infrared (IR) spectroscopy probes the vibrational modes of chemical bonds, with each functional group absorbing in a characteristic frequency range and leaving a diagnostic signature across the mid-IR region (4000-400 cm$^{-1}$) \cite{klein2013orgchem, griffiths2006spectra}. Among the available analytical techniques, IR is distinguished by its accessibility — instruments are inexpensive, sample preparation is minimal, measurements are non-destructive, and acquisition is rapid. These properties make it uniquely suited to resource-limited laboratories, high-throughput screening pipelines, and field deployment contexts. It is therefore a technique of particular interest to push toward its informational limits.

However, the structural information latent in an IR spectrum is only partially accessible by conventional means. The absorption band region (4000--1500 cm$^{-1}$) yields comparatively straightforward assignments: carbonyl stretches near 1700 cm$^{-1}$, broad hydroxyl stretches above 3000 cm$^{-1}$, and sharp amine signals are legible to a trained analyst and tractable to rule-based systems \cite{coates2000interpretation,larkin2017infrared}. The fingerprint region (1500–400 cm$^{-1}$) is more difficult, consisting of dense, overlapping absorptions arising from coupling skeletal vibrations, ring deformation modes, and bending interactions that encode detailed molecular information. This overlap renders the region largely intractable to manual interpretation \cite{alberts2024leveraging,larkin2017infrared}. Traditional computer-assisted structure elucidation (CASE) systems such as ACD/Structure Elucidator \cite{elyashberg2021case} partly addressed this gap by combining peak-matching against curated databases with rule-based fragment assembly, but their performance is fundamentally bounded by database coverage and by the manual effort required to encode interpretation rules. As a result, chemists have historically extracted only a small fraction of the structural information in IR spectra. The rise of data-driven methods fundamentally reframes this limitation: rather than asking how to encode rules for spectral interpretation, the question becomes how to learn them directly from data with deep learning (DL), a branch of artificial intelligence (AI).

Early DL approaches to IR analysis treated the problem as multi-label classification, training models to predict which functional groups are present in the molecule from its spectrum. Convolutional neural networks (CNNs) became the dominant choice in this regime, achieving high F1 scores across a range of standard functional group taxonomies \cite{fine2020funcgroups, enders2021funcgroups, jung2023automatic}. These methods established that the fingerprint region is richly informative when interpreted by sufficiently expressive models, but functional group classification, while useful for rapid characterization, falls short of the structural detail needed for unambiguous compound identification. Two molecules with identical functional group inventories can have very different connectivity, ring topologies, and substitution patterns, and these distinctions are precisely what determine reactivity, biological activity, and identity \cite{bemis1996molecules}.

Direct prediction of full molecular structure from spectra is a substantially harder task. In recent years, the use of transformer architectures, originally developed for natural language \cite{vaswani2017attention}, in structure elucidation tasks has yielded promising results \cite{alberts2024leveraging,wu2025patchbased, alberts2025setting}. These approaches commonly adapt the Vision Transformer (ViT) \cite{dosovitskiy2021vit}, encoding the spectrum as a sequence of patches and learning mappings from IR spectra and chemical formulae to structures represented as SMILES strings \cite{weininger1988smiles}. Evolutionary algorithms have also shown promise, particularly thus far for NMR-to-structure \cite{jin2025nmrsolver}, while multimodal approaches combining IR with NMR and mass spectrometry have demonstrated that spectroscopic modalities are complementary, with combined-modality models approaching expert-chemist accuracy \cite{alberts2024multimodaldataset, priessner2026multimodal}. Beyond single end-to-end models, Doh \textit{et al.} complement the generation workflow with specialized large language model sub-agents that perform post-hoc refinement of predicted structures \cite{noh2025iragent}; the ceiling of such refinement is nonetheless set by the underlying generative model — precisely the component this work seeks to characterize and improve.

Despite this progress, virtually every published IR-to-structure model shares a critical, often understated assumption: the molecular formula is provided as input alongside the spectrum at both training and inference time. This converts the structure elucidation problem from genuine de novo structure prediction into a substantially easier task: ranking candidate isomers consistent with a known molecular formula. Furthermore, obtaining an exact molecular formula is in itself a challenging task that requires, for example, a combination of high resolution mass spectrometry, isotopic analysis, and combustion analysis. A model that provides formula at inference is, in effect, a post-processing module within a multi-instrument pipeline that has already eliminated most of the analytical uncertainty. Alberts \textit{et al.} collapse the decoder's search space to formula-consistent candidates \cite{alberts2025setting}, which may obscure what is learned from spectral information.

In this work, we introduce a formula-free architecture for de novo IR-to-SMILES structure elucidation that combines two complementary mechanisms: sparse expert specialization with non-additive aggregation, and contrastive cross-modal alignment. The first is a novel enhancement of the popular Mixture-of-Experts (MoE) neural network layer \cite{shazeer2017moe,fedus2022switchtransformer} for transformer decoders, in which expert outputs are aggregated through fuzzy data-fusion operators — specifically our Linear Order Statistic Neuron (LOSN) \cite{veal2020losn} and our Choquet Integral MLP (ChIMP) \cite{islam2020chimp} — rather than the conventional Top-K weighted linear sum. This formulation retains the established benefits of sparse MoE (learned expert specialization, adaptive capacity allocation) while introducing order-sensitivity and explicit modeling of inter-expert synergy and redundancy. The second primary contribution is a modification of the transformer training objective that leverages auxiliary chemical information as a supervisory signal rather than a constraining inference-time input. Previously explored for simulated DFT-IR spectra in Zhang \textit{et al.} \cite{zhang2025formulafree}, this method entails optimizing a multi-criteria loss that includes a contrastive objective \cite{oord2019infonce, radford2021clip, yu2022coca} aligning encoded IR spectral representations with encoded structure representations, encouraging a latent geometry that is organized by chemical identity and therefore more conducive to molecular disambiguation in the absence of formula constraints.

Together, these contributions serve to inject a degree of compositional reasoning into structure elucidation models that is otherwise offloaded to the hard formula constraint present in previous approaches. Through the alignment mechanism, the model is encouraged to internalize the relationship between spectral features and atomic composition rather than bypass it. Our new MoE decoder complements this by providing the representational capacity to handle the expanded structural diversity that formula-free generation necessarily entails: where a formula-conditioned decoder navigates a constrained isomer subspace, a formula-free decoder must partition a far larger and more heterogeneous chemical space, and the order-sensitive and interaction-aware aggregation of LOSN and ChIMP provides a more expressive inductive bias for that task than a uniform dense FFN. We also identify and characterize the use of this aggregation in the processing of transformer encoder layers. By leveraging non-additive fuzzy logic, the model can learn to flexibly incorporate the respective representation layers learned by the spectral encoder into the cross-attention mechanism of the generative decoder.

Beyond architectural novelty, this work also advances how IR structure elucidation models are evaluated: standard Top-K exact-match accuracy, while reproducible, conflates structurally informative near-misses with random failures and provides no diagnostic signal for systematic deficiencies. We therefore introduce a substructure-level evaluation methodology that decomposes model errors across fragment classes — including ring systems, chain lengths, and functional group substitution patterns — enabling principled analysis of where formula-free models succeed and fail relative to their formula-conditioned counterparts.

% ==============================================================
\section{Related Works}
\label{sec:related_works}

\paragraph{Contrastive Learning.} Aligning heterogeneous modalities is naturally framed as a contrastive problem: an encoder is trained to map corresponding pairs to nearby points in a shared latent space while separating non-corresponding pairs, typically via the InfoNCE objective \cite{oord2019infonce}. CLIP \cite{radford2021clip} demonstrated this at scale with jointly trained image and text encoders, and CoCa \cite{yu2022coca} extended the paradigm to encoder-decoder architectures by jointly optimizing a contrastive alignment term alongside an autoregressive captioning loss, showing that the contrastive signal organizes the encoder representation space by cross-modal identity. Analogous alignment has since been applied to molecular property prediction \cite{wang2022gnn} and vibrational spectroscopy \cite{rocabert2025vibraclip, mirza2026secs}. Most directly related to our setting, Zhang and Ha \cite{zhang2025formulafree} adapt CoCa to IR-to-SMILES prediction on simulated data, finding that the contrastive term improves both accuracy and latent organization. We adopt this formulation and additionally exploit the learned embedding space for inference-time retrieval as a complement to decoder generation.

\paragraph{Non-Additive Data Fusion.} Combining multiple information sources is most often done through a linear weighted sum, which cannot represent the synergistic, redundant, or compensatory relationships that frequently hold between sources. The Choquet fuzzy integral \cite{murofushi1991choquet} generalizes the weighted mean by parameterizing aggregation with a fuzzy measure that assigns a worth to every coalition of sources, capturing such interactions explicitly; the Linear Order Statistic (LOS), a variant of Ordered Weighted Averaging \cite{yager1988owa}, instead aggregates by rank position, recovering operators from minimum to maximum within a single parameterized family. Both depart fundamentally from the additive linear convex sum (LCS) used in current MoE layers. Despite their established use in multi-sensor fusion and multi-criteria decision making \cite{auephanwiriyakul2002multisensor, grabisch2008decade}, adoption in deep learning has been limited by non-trivial monotonicity constraints. We previously showed that the Choquet integral admits a multilayer network representation (ChIMP) \cite{islam2020chimp}, with a M{\"o}bius-increment reparameterization (iChIMP) that enforces monotonicity by construction and trains under standard stochastic gradient descent (SGD). We also introduced the Linear Order Statistic Neuron (LOSN) \cite{veal2020losn} for order-sensitive, perceptron-style fusion over sorted inputs. These operators are the foundation of the aggregation mechanisms we bring to the transformer.

\paragraph{Aggregation in Transformer Architectures.} Two distinct stages of the transformer combine information through aggregation, and both conventionally rely on linear weighted sums. The first is cross-layer fusion: the decoder's cross-attention is conditioned only on the final encoder layer, discarding intermediate representations even though work in neural machine translation has repeatedly shown that layers encode qualitatively distinct information; lower layers capturing local, surface features and deeper layers more abstract, globally contextualized structure \cite{shi2016doesnmt, peters2018deeprep}. Peters \textit{et al.} \cite{peters2018deeprep} showed that task-specific scalar mixing of all layers outperforms using the final layer alone, a benefit confirmed by subsequent learned-weighting schemes \cite{dou2018nmt, yang2022nmt}. For IR encoding this organization is chemically meaningful, as early layers are positioned to capture local absorption features, while deeper layers encode structural context, so discarding intermediate layers risks losing critical evidence in the congested fingerprint region. The second stage, if utilized, is the Mixture-of-Experts (MoE) layer \cite{jacobs1991moe}, in which a gating network routes inputs to specialized expert sub-networks. Sparse MoEs activate only a subset of experts per token to enable scaling \cite{shazeer2017moe, fedus2022switchtransformer}, while differentiable variants such as Soft MoE \cite{puigcerver2024softmoe} and SMEAR \cite{muqeeth2024smear} replace hard routing with weighted averages of tokens or merged expert parameters. Fusion schemes for both stages remain linear at the point of aggregation; although fuzzy operators have been shown to integrate into deep networks \cite{price2019fuzzylayers}, to our knowledge they have never been applied to transformer layer fusion or MoE expert aggregation. This work treats both as instances of the same aggregation problem and addresses them with the non-additive operators introduced above.

% ==============================================================
\section{Methods}
\label{sec:methods}

\begin{figure}[h!]
    \centering
    \includegraphics[width=1.0\textwidth]{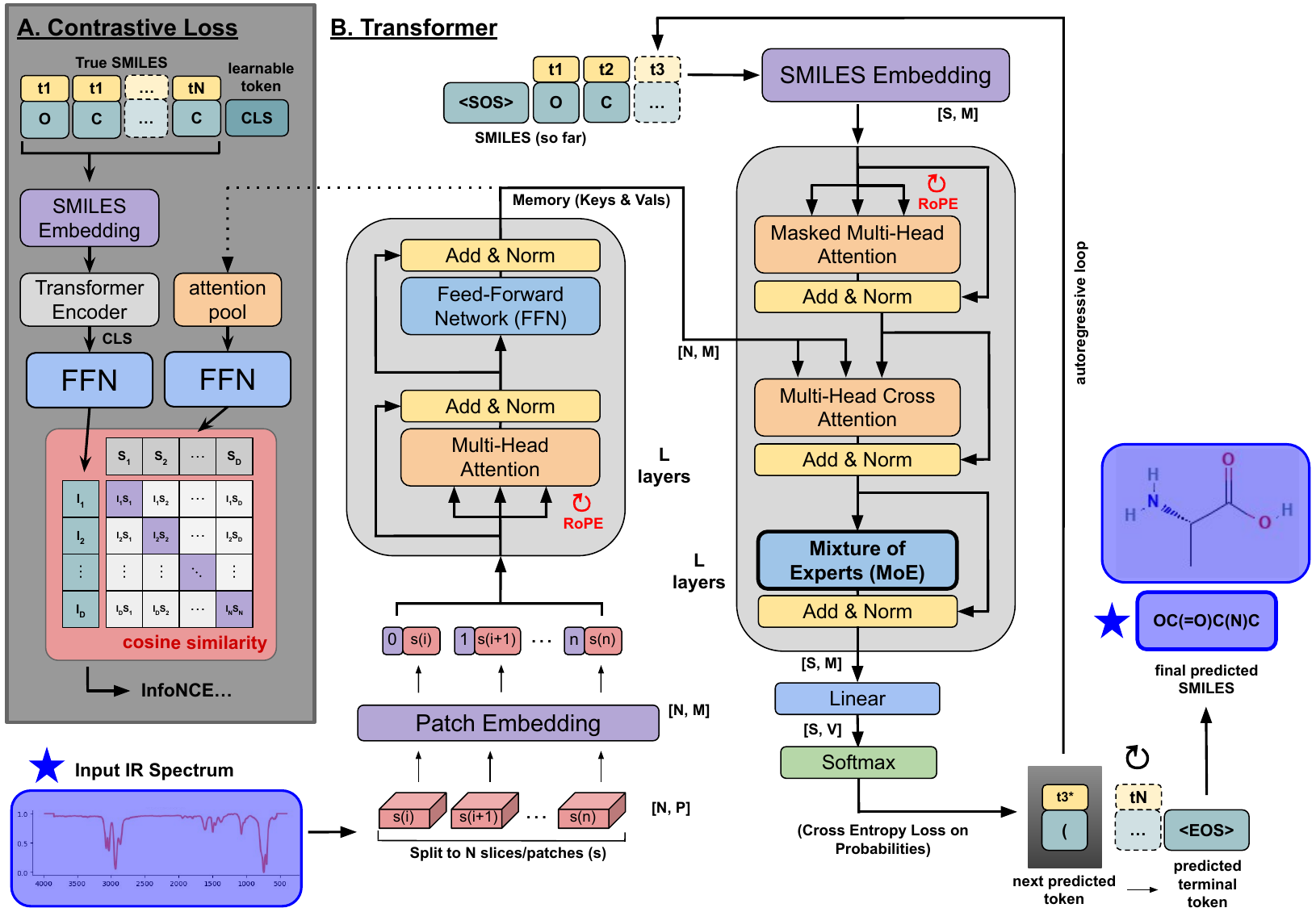}
    \caption{The proposed transformer architecture. In \textbf{(B)}, an input IR spectrum is first sliced into N patches of size P, which are then passed through L encoder layers, producing keys and values that are applied to the L decoder layers. A final linear layer and softmax produce the next token, which is either saved for backpropagation or passed back to the head of the decoder to continue the process (autoregressive inference). Within the decoder, the feed-forward network is replaced with an MoE layer (see Figure~\ref{fig:fmoe}). In \textbf{(A)}, the contrastive loss is derived via a cosine similarity matrix computed against latent representations of encoder outputs and molecular structures. Cross-entropy loss on decoder output probabilities is combined with this contrastive loss term to comprise the full optimization objective.}
    \label{fig:transformer}
\end{figure}

Our model architecture follows the encoder-decoder transformer paradigm \cite{vaswani2017attention}, in which an encoder maps an input sequence to a collection of continuous representations and an autoregressive decoder generates the output token-by-token conditioned on those representations. Both components are built from stacked layers of multi-head self-attention and position-wise feed-forward sub-layers with residual connections and post-layer normalization \cite{ba2016layernormalization}. Beam search generation yields a pool of $k$ ranked candidate SMILES strings $\mathcal{C}$ for input spectra. We depart from prior applications of the transformer in three primary respects, each detailed in the subsections below. Figure~\ref{fig:transformer}.B illustrates the full architecture; additional implementation details and hyperparameter settings are detailed in Appendix~\ref{supp:model_hparams}.

%% ─────────────────────────────────────────────────────────────
\subsection{Contrastive Cross-Modal Alignment}
\label{sec:alignment}

The primary training objective — cross-entropy over next-token predictions of the target SMILES string — provides gradient signal to the encoder only through the decoder's cross-attention, which is mediated by the decoder's own evolving representations. In the formula-free setting, where no compositional prior constrains the decoder's search space, this indirect supervision may be insufficient to produce a well-structured encoder latent space: two spectra corresponding to structurally distinct molecules may be mapped to nearby representations simply because their spectral profiles overlap in the fingerprint region. The decoder, lacking formula-based disambiguation, cannot correct for this at inference time. We address this by augmenting the primary objective with a contrastive alignment loss that directly supervises the geometry of the encoder's output space.

The alignment objective follows the CoCa framework \cite{yu2022coca}, instantiated with the CLIP-style InfoNCE loss \cite{radford2021clip,oord2019infonce}. This configuration was recently explored for this task in Zhang et. al \cite{zhang2025formulafree}, albeit for simulated data. An overview is visualized in Figure~\ref{fig:transformer}.A, which depicts the target SMILES string passing through a unimodal transformer encoder and the encoder memory undergoing attentional pooling, following the setup of \cite{yu2022coca}. For a batch of $N$ spectrum--molecule pairs $\{(s_i,m_i)\}_{i=1}^{N}$, spectral representations $\mathbf{z}_i^s$ are obtained by applying a multi-head attentional pooler over the encoder's final hidden states and projecting into a shared alignment space of dimension $d_{\text{align}}$ via a learned linear projection $f_{\phi}$:
\begin{equation}
    \mathbf{z}_i^s =
    f_\phi\!\left(\operatorname{AttnPool}\!\left(\mathbf{H}_i^{\text{enc}}\right)\right),
    \label{eq:spectral_proj}
\end{equation}
where $\mathbf{H}_i^{\text{enc}} \in \mathbb{R}^{P \times d}$ collects the encoder hidden states over $P$ spectral patch positions and $\operatorname{AttnPool}$ denotes cross-attention with a single learned query vector over those positions. Molecular representations $\mathbf{z}_i^{\text{smi}}$ are obtained from a dedicated unimodal SMILES encoder: the target SMILES tokens are embedded via the shared token embedding table, a learnable CLS token is appended to the sequence, and the result is processed by a stack of causal self-attention layers with no cross-attention to the spectrum. The CLS token's attention mask is fully unmasked so that it may attend to all preceding positions, giving it access to the complete sequence; its final hidden state is projected into the same alignment space via a learned linear $g_{\psi}$:
\begin{equation}
    \mathbf{z}_i^{\text{smi}} =
    g_\psi\!\left(\mathbf{h}_i^{\text{CLS}}\right),
    \label{eq:smiles_proj}
\end{equation}
where $\mathbf{h}_i^{\text{CLS}}$ is the unimodal encoder output at the CLS position. Critically, the unimodal SMILES encoder operates without access to the spectrum, ensuring that $\mathbf{z}_i^{\text{smi}}$ is a spectrum-agnostic structural representation; the contrastive loss then forces the spectrum encoder to produce representations that are geometrically consistent with these structure-derived embeddings. Both projections are trained end-to-end and $\ell_2$-normalized before similarity computation. The InfoNCE loss is computed symmetrically over the batch:
\begin{equation}
    \begin{aligned}
    \mathcal{L}_{\text{align}} =
    -\frac{1}{2N}\sum_{i=1}^{N}\Bigl[
    &\log\frac{\exp\!\left(\operatorname{sim}(\mathbf{z}_i^s,\,\mathbf
    {z}_i^{\text{smi}})/\tau\right)}
    {\sum_{j=1}^{N}\exp\!\left(\operatorname{sim}(\mathbf{z}_
    i^s,\,\mathbf{z}_j^{\text{smi}})/\tau\right)} \\
    +\,&\log\frac{\exp\!\left(\operatorname{sim}(\mathbf{z}_i^{\text{s
    mi}},\,\mathbf{z}_i^s)/\tau\right)}
    {\sum_{j=1}^{N}\exp\!\left(\operatorname{sim}(\mathbf{z}_
    i^{\text{smi}},\,\mathbf{z}_j^s)/\tau\right)}
    \Bigr],
    \end{aligned}
    \label{eq:infonce}
\end{equation}
where $\operatorname{sim}(\cdot,\cdot)$ denotes cosine similarity and $\tau$ is a learned temperature parameter, initialized to $1$ and exponentiated during training to remain strictly positive. The loss treats the $N-1$ remaining pairs in the batch as negatives for each anchor, so the effective difficulty of the contrastive task scales with batch size and, critically, with the structural diversity of the batch: batches containing molecules with similar structural motifs or overlapping spectral profiles present hard negatives that demand finer representational discrimination.

The alignment loss is active during training and validation but not at inference; the attentional pooler, unimodal SMILES encoder, and projection heads $f_{\phi}$ and $g_{\psi}$ do not participate in generation. The total training objective is a weighted combination of generative and alignment losses:
\begin{equation}
    \mathcal{L} =
    \mathcal{L}_{\text{CE}} + \lambda\,\mathcal{L}_{\text{align}},
    \label{eq:total_loss}
\end{equation}
where $\lambda$ is a scalar hyperparameter controlling the relative contribution of the alignment term. Section~\ref{sec:train_eval} provides further details on the compound loss objective.

%% ─────────────────────────────────────────────────────────────
\subsection{Encoder Layer Fusion}
\label{sec:encoder_fusion}

The flow of the transformer architecture, as shown in Figure~\ref{fig:transformer}.B, routes the output of the final encoder layer (memory) to the decoder's cross-attention module at every decoder layer $L$, discarding the intermediate representations produced by all preceding encoder layers. In this work, we seek to leverage these intermediate representations directly, taking inspiration from existing encoder fusion strategies and exploring the integration of fuzzy fusion operators. If we let $\{\mathbf{H}^{\ell}\}^{L}_{\ell=1}$ denote the hidden state sequences produced by the $L$ encoder layers, where $\mathbf{H}^{\ell} \in \mathbb{R}^{B \times T \times d_{model}}$ and $T$ is the number of spectral tokens, encoder fusion strategies receive the stacked tensor $\mathbf{X} \in \mathbb{R}^{B \times T \times L \times d_{model}}$ and produce a single fused memory $\hat{\mathbf{X}} \in \mathbb{R}^{B \times T \times d_{model}}$ that replaces $\mathbf{H}^{\ell}$ in the decoder's cross-attention. A shared gate network, consisting of a mean-pool over the layer dimension followed by a linear projection, produces per-layer logits that are softmax-normalized to gate probabilities $\mathbf{p} \in \mathbb{R}^{B \times T \times L}$. Traditionally, the $L$ layer representations are either averaged uniformly at each token position or gate probabilities $\mathbf{p}_{b,t} \in \Delta^{L-1}$ are used directly for input-dependent weighting. These are referred to as fixed and learned scalar mixing, respectively, as introduced in Peters \textit{et al.} \cite{peters2018deeprep}. The fuzzy fusion strategies explored in this work are detailed below; explicit details on each operator can be found in Appendix~\ref{supp:fuzzy_agg}.

\paragraph{LOSN.} Layer representations are ranked by their $\ell_2$-norm at each token position — a content-derived ordering that is independent of the gate — and aggregated via a shared learned rank-weight vector $\mathbf{w} \in \mathbb{R}^L$:
\begin{equation}
  \hat{\mathbf{X}}_{b,t} = \sum_{\ell=1}^{L} w_\ell \cdot \mathbf{H}^{\pi(\ell)}_{b,t}, \qquad \pi = \operatorname{argsort}\!\left(\|\mathbf{H}^\ell_{b,t}\|_2\right)_\downarrow,
  \label{eq:losn_fusion}
\end{equation}
where $\pi$ orders layers by descending representation norm at position $(b,t)$. The weight $w_1$ attaches to whichever layer produced the highest-norm representation at that position, $w_2$ to the second highest, and so on. The gate is computed, but not used for fusion; its cached values are retained for post-hoc comparison against the norm-based ordering. $\mathbf{w}$ is initialized with an exponential decay prior $w_\ell \propto \exp(\ell / L)$, encoding a soft inductive bias toward later layers that the model may override during training. In previous work, we formalized the extended linear order statistic (ELOS) \cite{kakula2020elos}, showing that the scalar rank-weight vector can be replaced with a per-dimension weight matrix $\mathbf{W} \in \mathbb{R}^{L \times d_{model}}$. This generalization is utilized for the LOSN in this work.

\paragraph{ChIMP.} The Choquet fusion strategy applies the $k=2$ additive restriction of ChIMP \cite{islam2020chimp, grabisch1997kadditive}, treating each encoder layer as a distinct information source and modeling pairwise interactions between layers through a learned fuzzy measure (FM). Layers are again ranked by their $\ell_2$-norm at each token position, and the fused representation is computed as:
\begin{equation}
  \hat{\mathbf{X}}_{b,t} = \sum_{\ell=1}^{L} \alpha_\ell \cdot \mathbf{H}^{\pi(\ell)}_{b,t} + \sum_{i < j} \beta_{ij} \cdot \min\!\left(\mathbf{H}^{\pi(i)}_{b,t},\, \mathbf{H}^{\pi(j)}_{b,t}\right),
  \label{eq:choquet_fusion}
\end{equation}
where $\min$ is applied element-wise and $\pi$ here orders layers by descending gate probability. The singleton terms $\mathbf{\alpha}_{\ell}$ capture the marginal contribution of the $\ell$-th ranked layer; the interaction terms $\mathbf{\beta}_{ij}$ model whether the $i$-th and $j$-th ranked layers are synergistic ($\mathbf{\beta}_{ij} > 0$) or redundant ($\mathbf{\beta}_{ij} < 0$) when combined. The FM is parameterized by $L + \binom{L}{2}$ free variables; for $L=6$ this is 21 parameters.

%% ─────────────────────────────────────────────────────────────
\subsection{Fuzzy Mixture-of-Experts (fMoE) Decoder}
\label{sec:fuzzy_decoder}

\begin{figure}[h!]
    \centering
    \includegraphics[width=1.0\textwidth]{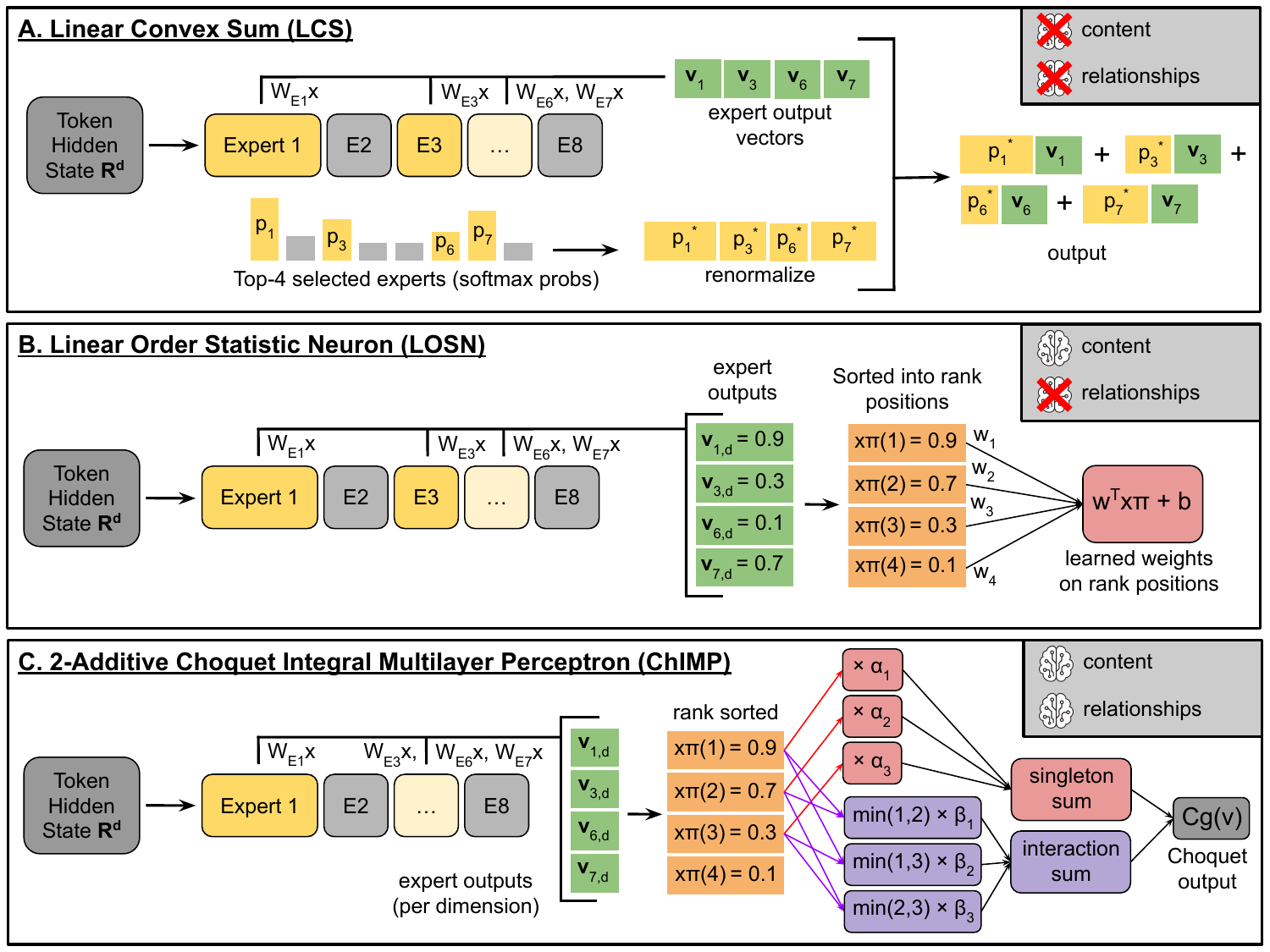}
    \caption{Standard and proposed expert aggregation methods for MoE. \textbf{(A)} The standard approach, with a sparse configuration. A router scores each expert for each token; only the top $k$ experts are used ($k=4$ is shown here). The router scores $p_e$ are then used directly as weights in a linear convex sum (LCS). \textbf{(B)} Our LOSN fMoE approach decouples aggregation from routing, sorting expert outputs by magnitude per output dimension and applying a learned vector to rank positions. This diagram depicts the per-dimension (ELOS) variation. \textbf{(C)} Our other fMoE approach, ChIMP models pairwise and higher-order interactions between experts with learned singleton and pairwise interaction weights.}
    \label{fig:fmoe}
\end{figure}

Standard transformer decoders consist of dense FFN networks applied at each layer after attention. In the formula-free setting, this architecture places the full burden of structural diversity on a single monolithic computation graph: the decoder must simultaneously handle aliphatic chains, heteroaromatic rings, halogenated substituents, and polar functional groups through the same learned weight matrix, relying on attention context alone to differentiate between them. We hypothesize that this is a poor inductive bias for formula-free generation, where the compositional diversity of the search space is strictly greater than in the formula-conditioned case. We therefore replace each decoder FFN sub-layer with a Mixture-of-Experts (MoE) layer \cite{jacobs1991moe}, composed of $E$ expert networks, and aggregate their outputs using fuzzy aggregation operators drawn from the fuzzy-set and information-fusion literature in addition to the linear weighted sums used in canonical MoE formulations. We refer to these novel MoE layers as fMoE aggregation. Figure~\ref{fig:fmoe} illustrates the specifications for the different MoE layer architectures.

\paragraph{Expert computation.} Each expert $e \in \{ 1,...,E\}$ is a position-wise FFN with parameters $\theta_e$ independent of all other experts. Given the decoder hidden state $\vx \in \R^{d_{model}}$ at a given position, each expert computes: 
\begin{equation} 
    \mathbf{v}_e = 
    \text{FFN}_e(\mathbf{x}, \theta_e), \quad e = 1, \ldots, E ,
    \label{eq:expert_output} 
\end{equation} 
yielding a set of $E$ candidate output vectors $\{\mathbf{v}_e\}$ of equal dimension. The aggregation problem is then: given these $E$ inputs, how should they be combined? In a sparse MoE, the answer is a weighted linear sum over router-assigned gates. This process is advantageous for scaling considerations (see Section~\ref{sec:related_works}), but maintains no effective strategy for fusion beyond the standard linear convex sum (LCS). The sparse MoE is shown in Figure~\ref{fig:fmoe}.A; a standard MoE follows a very similar process except all expert softmax probabilities contribute accordingly to the convex sum. To extend the potential of this configuration, we explore two operators wherein we decouple routing from aggregation and treat the process as a fuzzy information fusion problem. Again, we refer readers to Appendix~\ref{supp:fuzzy_agg} for additional context on fuzzy fusion operators.

\paragraph{MoE-LOSN.} Depicted in Figure~\ref{fig:fmoe}.B, MoE-LOSN modifies the decoder aggregation step by applying a separate LOSN (see Equation~\ref{eq:losn}) independently to each output dimension of the $E$ expert outputs, treating the expert index as the aggregation input: 
\begin{equation}
\begin{aligned}
    [\text{MoE-LOSN}(\mathbf{x})]_{d} &=
    f_{\text{LOSN}}\left([v_{1,d}, v_{2,d}, \ldots, v_{E,d}], \mathbf{w}^{(d)}\right), \\
    &\quad d = 1, \ldots, d_\text{model},
\end{aligned}
    \label{eq:moe-losn}
\end{equation} 
where $v_{e,d}$ is the $d$-th component of expert $e$'s output and $\mathbf{w}^{(d)}$ is the per-dimension weight vector. This preserves the LOSN's order-statistic semantics across expert outputs while remaining parallelizable across the output dimension.

\paragraph{MoE-ChIMP.} The Choquet Integral Multilayer Perceptron (ChIMP) \cite{islam2020chimp}  provides a strictly more expressive aggregation by modeling not only the marginal contribution of each expert but also pairwise and higher-order interactions between experts, through a learned fuzzy measure (FM) $g : 2^{X} \rightarrow  \R^{+}$. In this work, we apply the $k=2$ additive restriction \cite{grabisch1997kadditive}, retaining singleton and pairwise interaction terms while discarding higher-order coalitions to reduce the intractable scaling otherwise introduced at higher expert counts. Applied per output dimension, ChIMP aggregates expert outputs with explicit awareness of pairwise expert interactions, which is unavailable to any linear or order-statistic aggregation. The explicit MoE-ChIMP formulation, shown in Figure~\ref{fig:fmoe}.C, is defined as:
\begin{equation}
  \begin{aligned}
  [\text{MoE-ChIMP}(\mathbf{x})]_{d} &=
  C_g^{(2)}\left(\mathbf{h}^{(d)}; \boldsymbol{\alpha}^{(d)}, \boldsymbol{\beta}^{(d)}\right), \
  &\quad \mathbf{h}^{(d)} =
  [v_{1,d}, v_{2,d}, \ldots, v_{E,d}], \quad d = 1, \ldots, d_\text{model},
  \end{aligned}
  \label{eq:moe-chimp}
\end{equation}
where $\textbf{h}^{(d)} \in \mathbb{R}^{E}$ collects the $d$-th scalar output of each expert (i.e., $h_e = v_{e,d}$ in the notation of Eq.~\ref{eq:ichimp}) and $\alpha^{(d)} \in \mathbb{R}^E$ and $\beta^{(d)} \in \mathbb{R}^{\binom{E}{2}}$ are the learned singleton and pairwise interaction weights of the 2-additive FM for dimension $d$.

\paragraph{Load balancing.} The sparse Top-K LCS baseline (Figure~\ref{fig:fmoe}.A) is trained with the standard auxiliary load-balancing loss \cite{fedus2022switchtransformer}, which prevents the router from collapsing onto a small subset of experts and starving the remainder of training signal. This term regularizes the \emph{routing} stage and is specific to router-selected aggregation: because MoE-LOSN and MoE-ChIMP aggregate over all $E$ expert outputs by content-derived rank rather than over a router-selected subset, they are subject to no comparable collapse pressure and are trained without it. We give the formal definition in Appendix~\ref{supp:fuzzy_agg}. This asymmetry should be taken into account when interpreting the expert-utilization comparison in Appendix~\ref{supp:moe}, which is consequently not a controlled one.

% ==============================================================
\section{Experimental Setup}
\label{sec:experiments}

%% ─────────────────────────────────────────────────────────────
\subsection{Datasets}
\label{sec:datasets}

To set up a pipeline for supervised learning in the context of this problem, several considerations were taken into account concerning the data. In the interest of real-world application, and regarding data availability, it is advantageous to utilize simulated IR data as a component of this process, and as such a large simulated dataset of spectra synthesized via molecular dynamics and the PCFF forcefield first introduced by Alberts \textit{et al.} \cite{alberts2024leveraging} is used. This simulated dataset is paired with corresponding structures sampled from PubChem,  and is herein referred to as the IBM-MD dataset. Additionally, the National Institute of Standards of Technology (NIST) Chemistry WebBook \cite{wallace2024nist} provides experimentally-derived spectra and associated molecules; we synthesize a dataset from provided samples that includes only molecules containing C, H, O, N, S, P, and halogen atoms, and a heavy atom count of 6-13 similar to the subset explored in previous work and observed for the simulated IBM-MD dataset. We also benchmark performance on a third dataset synthesized via quantum chemical calculations by Zou \textit{et al.} \cite{zou2023qm9s}, called the QM9S dataset. This dataset contains 127,468 samples, however molecular diversity is limited to a maximum of 9 heavy atoms per molecule, and only H, C, N, O, or F atoms.

There are several preprocessing considerations employed to both synchronize the data language across the two competing datasets and constrain the feature space to dimensionality more amenable to initial experimentation. Roughly 1,500 real samples are duplicates of simulated samples, so these samples are first removed from the simulated dataset for pre-training. This ultimately results in 3,889 experimental samples and 633,864 simulated samples. IR spectra in both datasets are unified to a resolution of 2 cm$^{-1}$, yielding 1,791 evenly spaced absorption measurements from 400-3982 cm$^{-1}$. Min-max normalization is employed to standardize inputs in a manner conducive to machine learning models. 

It is often reasonable to represent molecular structure in graph form, but this is not easily utilized in common transformer models, particularly ViT. Luckily, advances in computational chemistry have led to the creation of a system for representing molecules using ASCII characters, known as the Simplified Molecular Input Line Entry System (SMILES) \cite{weininger1988smiles}. SMILES strings are built via a depth-first tree traversal of a standard chemical graph, and they are Hydrogen-implicit. They are fundamentally a one-to-many mapping, with a canonicalization protocol to unify the algorithm's application. SMILES strings can be tokenized, or converted to numerical representations, in various ways; the method employed here is an atom-wise approach using a regular expression (RegEx) introduced in Schwaller \textit{et al.} \cite{schwaller2019regex}. For the chemical subspace explored in this work, this results in a SMILES vocabulary size of 47.

Data augmentation is often beneficial for training machine learning models, and it has been shown to be effective for transformer-based structure elucidation \cite{alberts2024leveraging,wu2025patchbased}. We find it best to employ SMILES enumeration \cite{bjerrum2017smilesenum}, a data augmentation technique that exploits the existence of non-canonical SMILES strings to expand the training corpus. Alberts \textit{et al.} find that augmenting each training sample with two non-canonical SMILES representations boosts Top-1 accuracy by around 5\% \cite{alberts2025setting}; we explore varying degrees of SMILES augmentation to further characterize the benefit of this approach. Results of this analysis are detailed in Appendix~\ref{supp:smiles_aug}.

\subsection{Training and Evaluation}
\label{sec:train_eval}

\paragraph{Training Objective.} The model is trained end-to-end to autoregressively generate SMILES token sequences from IR spectral input. The primary training signal is the standard cross-entropy loss over the SMILES vocabulary, which at each decoding step penalizes the model's predicted token distribution against the ground-truth next token under teacher forcing:
\begin{equation}
    \mathcal{L}_{\mathrm{CE}} = 
    -\frac{1}{N} \sum_{n=1}^{N} \sum_{t=1}^{T_n} \log p_\theta\left(y_t^{(n)} \mid y_{<t}^{(n)}, \mathbf{X}^{(n)}\right),
    \label{eq:ce}
\end{equation}
where $N$ is the number of training examples, $T_n$ is the target sequence length for the $n$-th example, $y_{t}^{(n)}$ is the ground truth token at position $t$, and $\mathbf{X}^{(n)}$ is the encoded spectral representation. As defined in Equation~\ref{eq:total_loss}, $\Loss_{\text{CE}}$ is combined with a $\lambda$-weighted contrastive loss term $\Loss_{\text{align}}$ to form the full optimization signal. The relative contribution of $\Loss_{\text{align}}$ to the total training signal is a tunable hyperparameter; Zhang \textit{et al.} find that raising this value leads to better structuring of latent embeddings \cite{zhang2025formulafree}. Regardless, the cross entropy loss term is the primary signal driving the whole architecture to learn how to generate structures. Specific training parameter settings, such as the learning rate and number of epochs, are covered in Appendix~\ref{supp:training}.

\paragraph{Decoding and Generation.} At inference time, the model generates candidate SMILES strings $\mathcal{C}$ via beam search with beam width 10, retaining the Top-K sequences by cumulative log-probability. No molecular formula is provided to the decoder at any point during inference. Each generated sequence is then parsed and canonicalized using RDKit \cite{rdkit}, and then compared against the ground-truth structure using the equivalence of InChI \cite{heller2013InChI}. Invalid SMILES strings that fail valence or sanitation checks are discarded from the candidate set prior to evaluation. Decoding is deterministic (no sampling, temperature $1.0$) with a maximum generation length of 100 tokens.

\paragraph{Evaluation Metrics.} Model performance is evaluated at multiple levels of structural resolution. The primary metric is Top-K exact-match prediction accuracy for $K \in \{1,5,10 \}$. A prediction is counted as correct at rank $k$ if the ground truth appears among the Top-K beam candidates. For predictions that do not exactly match the ground truth, Tanimoto similarity \cite{bajusz2015tanimoto} between chemical fingerprints provides a continuous measure of structural proximity; we report the distribution of Tanimoto similarities over incorrect Top-1 predictions for different fingerprint representations.

While exact match and Tanimoto similarity are both descriptive molecule-level metrics that are widely adopted in the literature, they can obscure qualitatively different error modes when evaluating the performance of these types of models. For instance, a model may recover the correct carbon skeleton while misclassifying a single functional group, or confuse structurally similar ring systems in ways invisible to fingerprint-based similarity. We therefore additionally report a substructure-level evaluation, decomposing predictions and ground truths into different fragment vocabularies and computing precision and recall over the resulting substructure sets. This provides finer-grained diagnostic signal regarding where in chemical space the model succeeds and fails, and is particularly informative for characterizing errors as a function of molecular complexity.

When CoCa is active, we also evaluate alignment quality through a spectrum-to-SMILES retrieval task defined entirely within the shared CoCa latent space. This method was previously explored for contrastive alignment of vibrational spectroscopic modalities in Rocabert-Oriols \textit{et al.} \cite{rocabert2025vibraclip}; we include it here as a diagnostic orthogonal to generation accuracy. For all $N$ test examples, the L2-normalized spectrum embeddings $\mathbf{S} \in \mathbb{R}^{N \times d_{\text{align}}}$ and SMILES embeddings $\mathbf{M} \in \mathbb{R}^{N \times d_{\text{align}}}$ are assembled into a full $N \times N$ cosine similarity matrix $\mathbf{A} = \mathbf{S}\mathbf{M}^\top$. The diagonal of $\mathbf{A}$ gives the per-pair cosine similarity; its mean quantifies how tightly matched pairs are drawn together on the unit hypersphere relative to all cross-pair distances. For each query spectrum $i$, the rank of its ground-truth SMILES is defined as the number of gallery entries at least as similar: $r_i = |{j : A_{ij} \geq A_{ii}}|$, with ties counted inclusively so that $r_i \geq 1$ always. Recall at $K$ (R@K) is the fraction of queries for which $r_i \leq K$, reported at $K \in {1, 5, 10}$. These metrics characterize whether the CoCa objective has induced a latent geometry in which a query spectrum, presented with the full test set as its gallery, reliably recovers its paired SMILES among the nearest neighbors.

% ==============================================================
\section{Results}
\label{sec:results}

This section details various experiments and the effect of different architectural modifications. We evaluate our model across three datasets: the IBM molecular-dynamics simulated dataset (IBM-MD), the QM9S quantum chemistry simulated dataset~\cite{zou2023qm9s}, and the NIST experimental dataset~\cite{wallace2024nist}. Unless otherwise stated, all models are trained and evaluated \emph{without} auxiliary chemical formula input. The primary metric, Top-K accuracy, provides the most direct measure of overall performance.

%% ─────────────────────────────────────────────────────────────
\subsection{IR Structure Elucidation}
\label{sec:results:main}

\begin{table}[htbp]
\centering
\caption{IR-only structure elucidation performance on NIST experimental (fine-tuned) and IBM-MD simulated spectra. Missing table entries for previous works exist because they were not explicitly reported.}
\label{tab:main_comparison}
\begin{tabular}{lllcccccc}
\toprule
& \multicolumn{2}{c}{\textbf{Model Input}} & 
    \multicolumn{3}{c}{\textbf{IBM-MD Simulated}} &
    \multicolumn{3}{c}{\textbf{NIST Experimental}} \\
\cmidrule(lr){2-3}\cmidrule(lr){4-6}\cmidrule(lr){7-9}
Model & IR & Formula &
  Top-1 & Top-5 & Top-10 &
  Top-1 & Top-5 & Top-10 \\
\midrule
%% ---- formula-conditioned references (not directly comparable) ----
Alberts et~al.~\cite{alberts2024leveraging}   & \cmark & \cmark & 36.43 & 62.49 & 70.00 & 44.39 & 66.85 & 69.79 \\
Alberts et~al.~\cite{alberts2025setting}      & \cmark & \cmark & 50.62 & --- & 72.39 & 63.25 & 79.15 & 83.56 \\
Wu et~al.~\cite{wu2025patchbased}             & \cmark & \cmark & 50.96 & 74.24 & 78.80 & 53.23 & 71.99 & 76.54 \\
\midrule
%% ---- IR-only baselines ----
Alberts et~al.~\cite{alberts2024leveraging}   & \cmark & \xmark & 17.01 & 33.60 & 39.37 & --- & --- & --- \\
Wu et~al.~\cite{wu2025patchbased}             & \cmark & \xmark & 23.84 & 42.24 & 50.17 & --- & --- & --- \\
%% ---- ours ----
\textbf{Ours} & \cmark & \xmark & \textbf{27.12} & \textbf{44.62} & \textbf{53.96} & \textbf{31.83} & \textbf{50.13} & \textbf{57.23} \\
\bottomrule
\end{tabular}
\end{table}

Table~\ref{tab:main_comparison} presents the primary comparison between our model and IR-only baselines on the NIST experimental dataset and the IBM-MD simulated dataset. Our best-performing model utilizes the MoE-LOSN decoder configuration and encoder layer fusion, and is trained with the CoCa contrastive loss configuration. Results analyzing these respective contribution components are detailed in the sections below.  For the NIST dataset, IBM-MD-trained models are fine-tuned on a training subset of experimental NIST data. In the literature, few results describe performance in the IR-only regime; benchmark metrics measure models which take the formula as input alongside the spectrum. These previously reported metrics are shown in the table for the IBM-MD dataset; to our knowledge, no previous IR-only model implementations have been tested specifically on the NIST experimental dataset. For the former, our model achieves state-of-the-art Top-K accuracy in the IR-only setting; we report an impressive 31.83\% Top-1 accuracy after fine-tuning on experimental data.

\begin{figure}[htbp]
  \centering
  \includegraphics[width=1.0\textwidth]{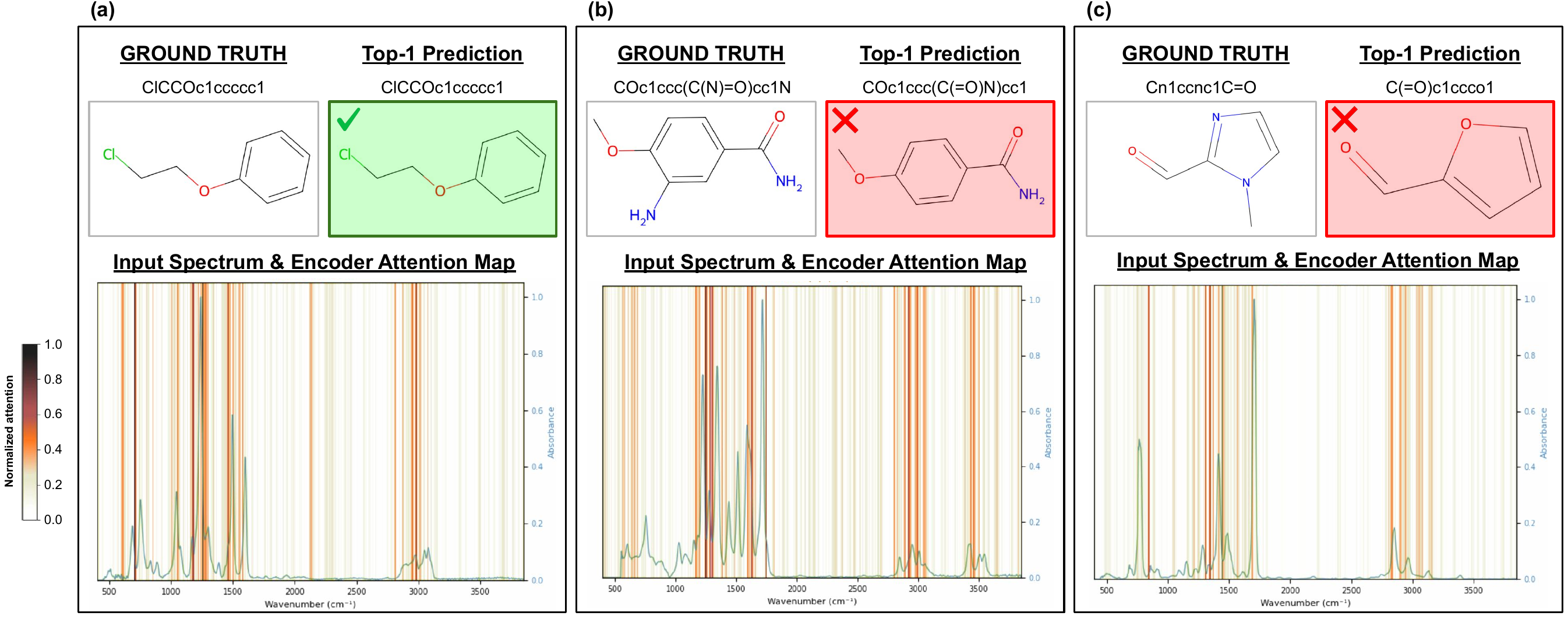}
  \caption{%
    Qualitative prediction examples on NIST experimental spectra. \textbf{(a)}~Correct Top-1 prediction with encoder cross-attention heat map. We see the model focuses around 650-750 cm$^{-1}$ for the alkyl chlorine, notes the aryl ether peak at $\sim$ 1220 cm$^{-1}$, as well as the two diagnostic peaks between 1450-1600 cm$^{-1}$ from carbon-carbon stretching within the aromatic ring. \textbf{(b)}~Near-miss (Tanimoto = 0.52, later correctly identified at beam \#8): The model misses a primary amine group, which is inherently harder to discern with the presence of overlapping methoxy and amide groups. \textbf{(c)}~Failure case (best-beam Tanimoto = 0.18): The model fails to identify the imidazole \ch{N-CH3} methyl group just under 3000 cm$^{-1}$. More attention is weighted to less distinguishing areas, and a plausible but dissimilar heteroaromatic ring system is predicted.
  }
  \label{fig:mol_preds_attention}
\end{figure}

In Figure~\ref{fig:mol_preds_attention}, we have representative prediction outcomes for experimental samples with our model: a correct Top-1 prediction, a near-miss with high Tanimoto similarity, and a failure case. Also shown is the corresponding input spectrum for each example with encoder cross-attention weights overlaid. This figure visualizes how the model processes input spectra, paying attention to specific wavenumber ranges during the generation of respective molecules. For many test samples, this is functionally analogous to how an expert chemist might analyze an IR spectrum. identifying diagnostic peaks and absorption behaviors. For failure cases, such as the one shown in Figure~\ref{fig:mol_preds_attention}.c, we often see the model either paying attention to the wrong ranges or conflating overlapping absorption bands. For this specific example, 1-methyl-1H-imidazole-2-carboxaldehyde is mistaken for 2-furaldehyde/furfural, which, despite the low Tanimoto similarity, is a highly plausible mistake given that both molecules are small, five-member heteroaromatic rings with an attached aldehyde group.

\begin{figure}[h!]
    \centering
    \includegraphics[width=1.0\textwidth]{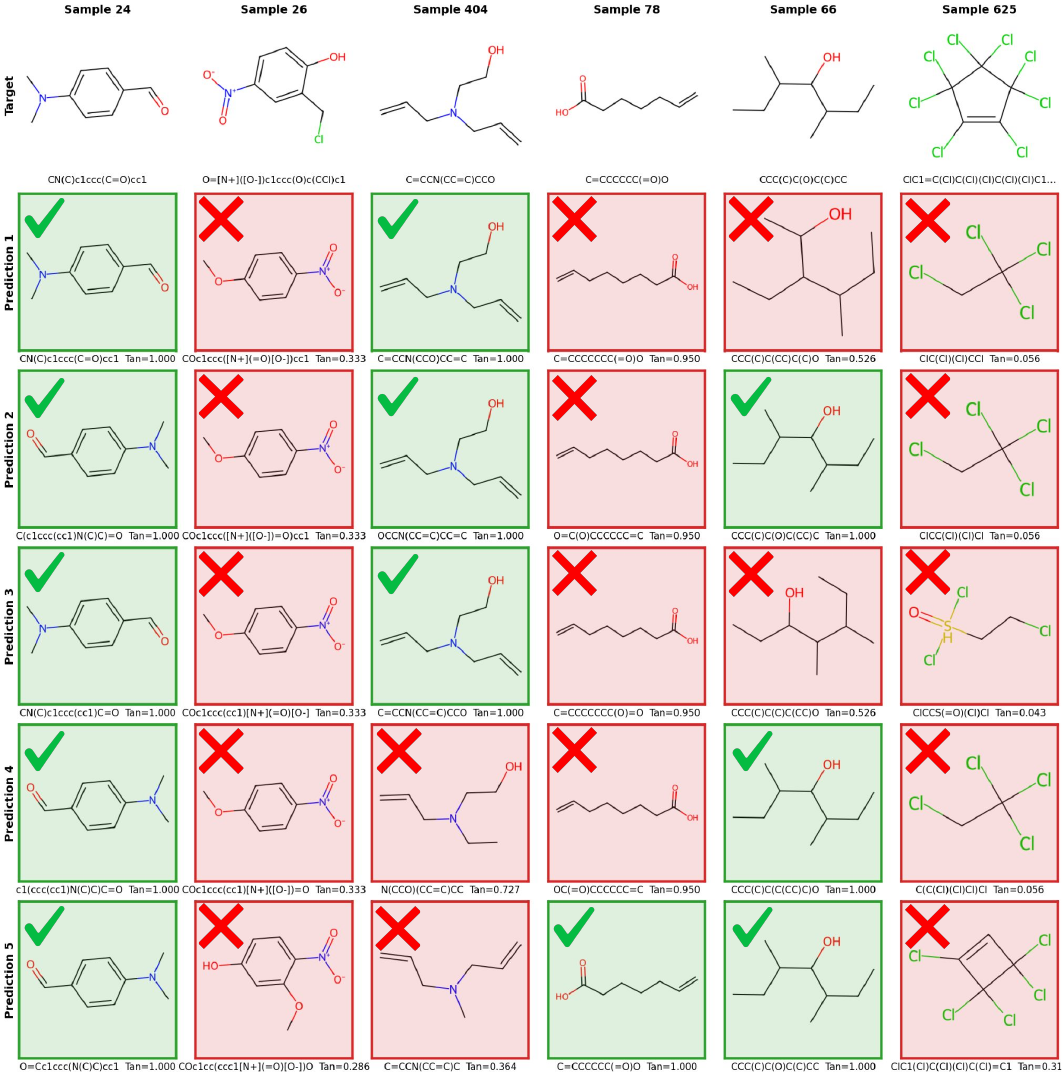}
    \caption{Select Top-5 predictions for the MoE-LOSN model on the NIST experimental test set. The model takes unseen experimental IR spectra from the held-out test dataset as input, generating a set of ordered predictions. These are then compared against ground truth molecules associated with the test spectra. Correct predictions, which span the chemically equivalent SMILES representation space for respective molecules, are highlighted in green and marked with checks, whereas incorrect predictions are highlighted in red. SMILES strings and Tanimoto similarity for each prediction are also shown.}
    \label{fig:mol_preds_top5}
\end{figure}

Visualization of select test samples and their Top-5 predictions are shown in Figure~\ref{fig:mol_preds_top5}. A range of test samples are chosen to better illustrate common failure modes in addition to general versatility. While each generated molecule is canonicalized before comparison against the ground truth, this plot visualizes the exact generated string with RDKit's 2D coordinate generation algorithm \cite{rdkit}. This algorithm is sensitive to the input atom order, and we can see that this sometimes results in inconsistent 2D visualizations for otherwise equivalent correct predictions (i.e., Sample 24). For Sample 78, the first four predictions contain eight carbons, while the fifth prediction contains seven carbons, consistent with the target. Of note is the general proximity of candidate molecules to true samples; for example, even Sample 26, which doesn't see an exact match in the Top 5, still has multiple correct functional groups consistently identified. Sample 625 is composed primarily of chlorine atoms; the presence of the latter is consistently identified, likely due to the presence of strong \ch{C-Cl} stretching vibrations between 600-800 cm$^{-1}$. Despite this, Octachlorocyclopentene is a larger molecule in the context of the dataset and this may contribute to the observed difficulty in matching structural components such as the cyclopentene ring scaffold or the correct number of chlorine atoms.

%% ─────────────────────────────────────────────────────────────
\subsection{Contrastive Loss}
\label{sec:results:contrastive}

\begin{figure}[h!]
  \centering
  \includegraphics[width=1.0\textwidth]{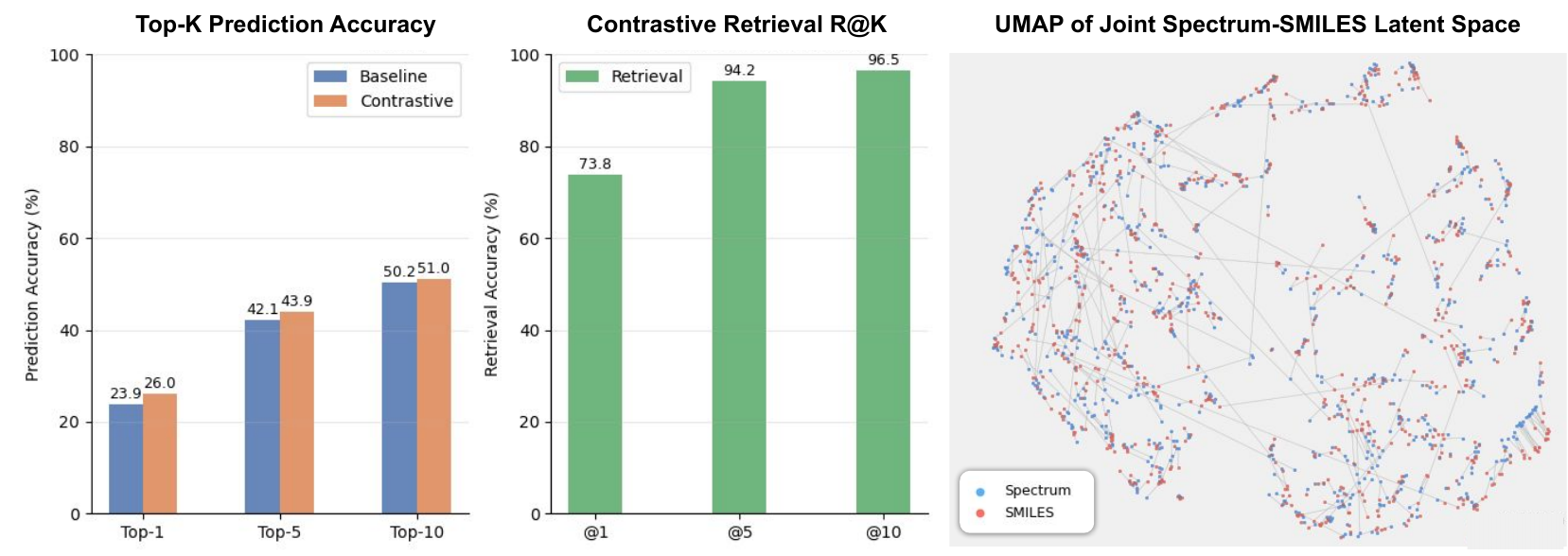}
  \caption{Effect of contrastive loss on model performance. (Left): Top-K prediction accuracy of our baseline model and the same model trained with the contrastive loss. (Middle): Retrieval accuracy R@K of the contrastive model at increasing candidate set sizes (R@1, R@5, R@10), showing the fraction of spectra whose correct molecule is recovered within the first R matches by embedding similarity. (Right): UMAP visualization of the joint embedding space aligned by the contrastive loss. Lines connect matched pairs.
  }
  \label{fig:contrastive}
\end{figure}

The effect of adding the contrastive loss $\Loss_{align}$ term to the standard cross-entropy loss $\Loss_{CE}$ is detailed in Figure~\ref{fig:contrastive}. We weight the contribution of $\Loss_{align}$ to the total objective with $\lambda=1.0$, and utilize a 2-layer unimodal SMILES transformer with $d_{model}=512$. Doing this improves Top-K accuracy, primarily in the most restrictive case (Top-1). Findings for these metrics are in line with those observed in Zhang \textit{et al.} \cite{zhang2025formulafree} on simulated data. We also compute additional metrics and include a unified UMAP visualization to better characterize the effect of the contrastive signal. At test time, both spectra and SMILES are passed through their respective encoders and evaluated for retrieval accuracy (see Section~\ref{sec:train_eval} for specifics). We report high R@K retrieval accuracy, even when considering only a single candidate. This suggests that the model has learned a well-aligned latent space where spectra and SMILES pairs are highly proximate. This is further validated by the UMAP visualization, which shows tight clustering with few outliers. The mean cosine similarity between matched pairs is also high, at $83.44$.

%% ─────────────────────────────────────────────────────────────
\subsection{Encoder Layer Fusion}
\label{sec:results:encoder_fusion}

\begin{table}[htbp]
\centering
\caption{%
  Top-K accuracy of different encoder layer fusion strategies. Results convey performance on experimental NIST data after fine-tuning respective IBM-MD trained models. All models use the contrastive loss as well.
}
\label{tab:encoder_fusion}
\begin{tabular}{lcccc}
\toprule
Model & Top-1 & Top-5 & Top-10  & Additional Parameters \\
\midrule
None          & 26.01 & 43.89 & 51.20 & --- \\
Simple mean   & 22.49 & 39.66 & 46.16 & --- \\
Gate-weighted & 25.55 & 42.59 & 50.65 & $L(d_{model}+1)$ \\
LOSN          & \textbf{27.61} & 45.14 & 51.49 & $L(d_{model}+1)$ \\
ChIMP         & 27.10 & \textbf{45.71} & \textbf{51.62} & $L(d_{model}+1) + \binom{L}{2}$ \\
\bottomrule
\end{tabular}
\end{table}

As detailed in Section~\ref{sec:related_works}, leveraging the information of all encoder layers can be advantageous to transformers. Table~\ref{tab:encoder_fusion} details an ablation of fusion strategies on the contrastive-trained model from Section~\ref{sec:results:contrastive}. We can see that the simple mean formation, where we have a uniform average over layers, degrades performance. This is not surprising, as it necessitates equal contribution from all layers with no room for dynamic consideration. Interestingly, the addition of dynamic weighting with the gate-weighted strategy doesn't have much of an effect. This is in contrast to the fuzzy fusion operator strategies, which each improve performance significantly. The rationale behind the gate-weighted performance may come down to the gate itself; training imbalance and gradient interference from the convoluted optimization path may blur the effect of this small component. With LOSN and ChIMP, there is no reliance on a well-calibrated gate — they instead compute aggregation on a per-token basis leveraging the content-derived $\ell_2$-norm sort. While the fusion methods do add learned parameters, the additions are negligible relative to the model as a whole, suggesting that the observed performance differences are purely attributable to the inductive bias of the aggregation operator.

%% ─────────────────────────────────────────────────────────────
\subsection{Decoder Aggregation Ablation}
\label{sec:results:decoder}

\begin{table}[htbp]
\centering
\caption{%
  Top-K accuracy of different decoder aggregation strategies for various datasets. The IBM-MD and QM9S results are on held-out simulated samples, while the NIST results convey performance on experimental data after fine-tuning IBM-MD trained models.
}
\label{tab:decoder_agg}
\begin{tabular}{lccccccccc}
\toprule
& \multicolumn{3}{c}{\textbf{IBM-MD Simulated}} &
  \multicolumn{3}{c}{\textbf{QM9S Simulated}} &
  \multicolumn{3}{c}{\textbf{NIST Experimental}} \\
\cmidrule(lr){2-4}\cmidrule(lr){5-7}\cmidrule(lr){8-10}
Model & Top-1 & Top-5 & Top-10 & Top-1 & Top-5 & Top-10 & Top-1 & Top-5 & Top-10\\
\midrule
Dense FFN          & 24.21 & 42.99 & 50.23 & 52.15 & 69.53 & 71.48 & 27.21 & 45.14 & 51.49 \\
MoE-LCS            & 25.06 & 42.63 & 50.76 & 55.58 & 72.40 & 74.98 & 27.86 & 45.84 & 51.52 \\
MoE-LOSN           & \textbf{27.12} & \textbf{44.62} & \textbf{53.96} & \textbf{63.28} & \textbf{86.13} & \textbf{89.06} & \textbf{31.83} & \textbf{50.13} & \textbf{57.23} \\
MoE-ChIMP          & 25.51 & 41.77 & 49.85 & 58.01 & 76.66 & 80.22 & 28.86 & 47.99 & 54.93 \\
\bottomrule
\end{tabular}
\end{table}

Table \ref{tab:decoder_agg} details model performance for different decoder configurations. All models utilize the best contrastive loss configuration from Section~\ref{sec:results:contrastive} and LOSN encoder fusion per Section~\ref{sec:results:encoder_fusion}. MoE models are sparse, consisting of 8 experts with $k=4$ as shown in Figure~\ref{fig:fmoe}. The total active parameters of these 4 experts match the parameter count of the dense FFN for consistency. Additional hyperparameter settings are detailed in Appendix~\ref{supp:model_hparams}.

While the MoE-LCS model slightly outperforms the standard dense (FFN) approach across datasets, fuzzy aggregation methods show a distinct performance advantage. The MoE-LOSN model achieves the highest Top-1 accuracy of all tested models on the experimental test set. The MoE-ChIMP approach is only slightly better than the MoE-LCS baseline, which may be due to the fact that the Choquet formulation is less advantageous for Mixture of Experts due to the inherent lack of distinct heterogenity among experts. Interestingly, all models perform significantly better on QM9S samples; as suggested in Wu \textit{et al.} \cite{wu2025patchbased}, this is likely due to the relative simplicity of spectra in this dataset. Density Functional Theory (DFT) is inherently static, wheras Molecular Dynamics (MD) simulations can capture mode coupling, fermi resonances, and anharmonicities naturally present in real-world measured spectra \cite{gastlegger2017moleculardynamics}. 

%% ─────────────────────────────────────────────────────────────
\subsection{Tanimoto Similarity and Structural Proximity}
\label{sec:results:tanimoto}

Beyond exact-match accuracy, we report Tanimoto similarity to characterize structural proximity of incorrect predictions. Table~\ref{tab:tanimoto} shows that even when our model does not recover the exact structure, its predictions remain chemically proximate to the target. Structural proximity is also further characterized by measuring the deviation in molecular weight between ground truth and predicted molecules. Finally, the model's ability to generate structures with the correct chemical formulae is measured.

\begin{table}[htbp]
\centering
\caption{%
  Structural proximity of Top-1 predictions on NIST experimental dataset for models with and without formula input. Molecular weight deviation (MAE) Tanimoto similarity for different fingerprint strategies, and formula accuracy (exact atom-count match) are reported.
}
\label{tab:tanimoto}
\begin{tabular}{lllcccccc}
\toprule
& \multicolumn{2}{c}{\textbf{Model Input}} &
\multicolumn{3}{c}{\textbf{Tanimoto Similarity}} &
  \multicolumn{2}{c}{\textbf{Compositional Similarity}} \\
\cmidrule(lr){2-3}\cmidrule(lr){4-6}\cmidrule(lr){7-8}
Model & IR & Formula &
  MACCS & RDKit & Morgan &
  MW MAE & Formula Acc. \\
\midrule
Dense FFN & \cmark & \xmark & 0.754 & 0.603 & 0.558 & 14.36 & 38.96 \\
Dense FFN & \cmark & \cmark & 0.797 & 0.636 & 0.595 & 2.94 & 65.62 \\
MoE-LOSN  & \cmark & \xmark & 0.762 & 0.612 & 0.575 & 13.29 & 45.11 \\
MoE-LOSN  & \cmark & \cmark & 0.831 & 0.649 & 0.621 & 2.21 & 74.43 \\
\bottomrule
\end{tabular}
\end{table}

%% ─────────────────────────────────────────────────────────────
\subsection{Substructure-Level Error Analysis}
\label{sec:results:substructure}

\begin{figure}[h!]
    \centering
    \includegraphics[width=1.0\textwidth]{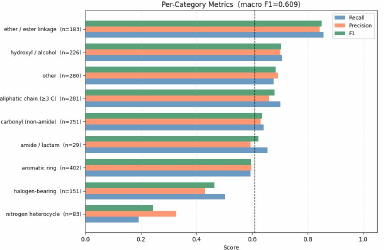}
    \caption{Per-fragment-class precision, recall, and F1 scores for BRICS fragment classes on held-out NIST experimental spectra (Top-1 prediction). Fragments are classified into 9 chemically distinct named categories as shown here; fragments not fitting into any of the established categories are classified as "other." The frequency of each fragment class across the test dataset, n, is recorded next to each class name. The vertical dashed line marks the macro F1 score for all classes.}
    \label{fig:brics}
\end{figure}

Exact-match accuracy does not capture partial structural recovery. We therefore supplement Top-K metrics with various substructure metrics and similarity characterization. Figure~\ref{fig:brics} reports fragment-level precision and recall across chemically significant fragment classes through Breaking of Retrosynthetically Interesting Chemical Substructures (BRICS) decomposition \cite{degen2008brics}, providing finer-grained insight into where errors concentrate. More information on the BRICS fragmentation process, specifically the custom classes shown in Figure~\ref{fig:brics}, can be found in Appendix~\ref{supp:fragments_brics}. 

Also recorded is the relative abundance of BRICS classes across the test dataset. This spread is similar to the spread of the training dataset, and is not necessarily prohibitive in that some classes are simply more chemically common, a reality reflected by the distribution of existing data sources. We see that certain BRICS fragments have lower performance, and this is chemically meaningful: for example, ether/ester linkages are spectrally loud (i.e., carbonyl stretches), whereas halogens are more quiet and nitrogen heterocycles have complex ring structures and substitution patterns that affect absorption considerably.

\begin{figure}[h!]
    \centering
    \includegraphics[width=1.0\textwidth]{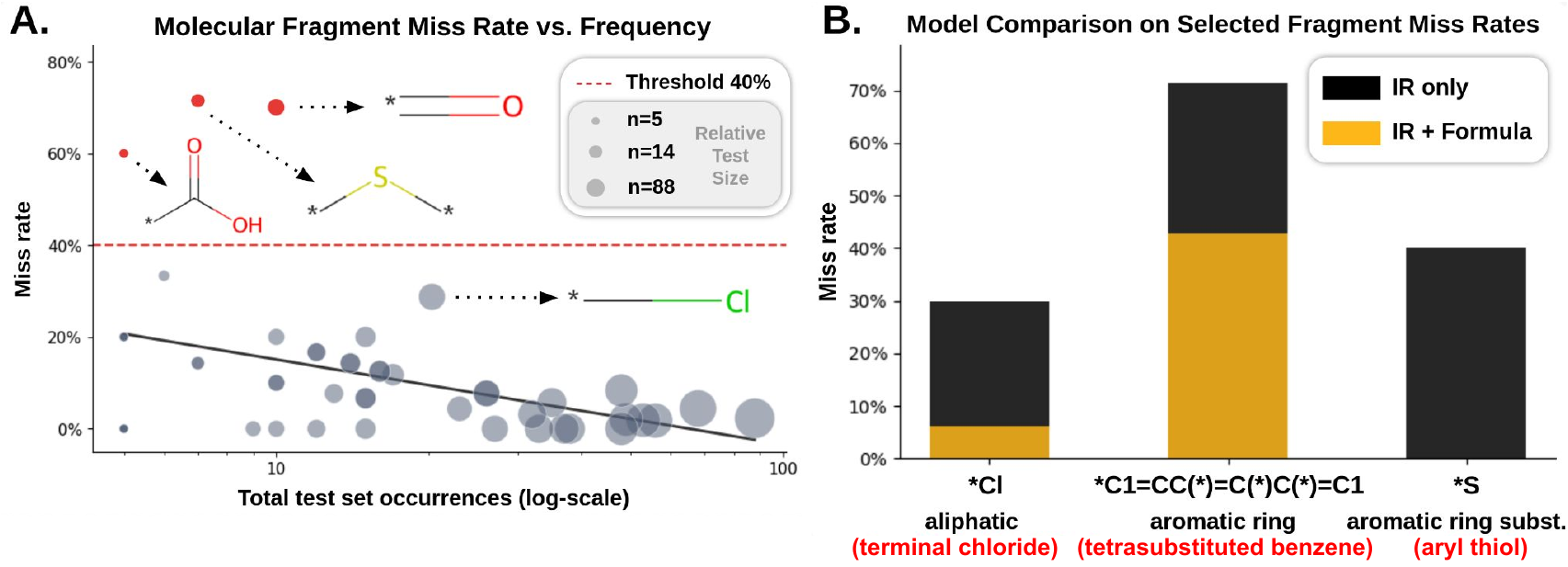}
    \caption{Substructure analysis. (A): Miss rate of sp\textsuperscript{3}-cut fragments vs. their frequency in the held-out test set. The SMILES of select outlier fragments are overlaid on the plot. (B): A comparison of select fragment miss rates for the standard IR-only model and a variation where the formula is also provided as input. Labeled is the SMILES string corresponding to the shown fragment, as well as respective sp\textsuperscript{3}-cut fragment classes and taxonomic context.}
    \label{fig:fragments}
\end{figure}

Aside from BRICS, we also evaluate performance using a custom fragmentation engine. This method differs from BRICS in that it is not retrosynthetically motivated, and rather cuts where sp\textsuperscript{3} carbons bridge structural motifs such as aromatic rings. By separating core chemical structures from their modification patterns, we obtain a richer characterization of structural motifs that is distinct from bond environment rule-based approaches (such as BRICS). For more information on this fragmentation engine, refer to Appendix~\ref{supp:fragments_custom}.

Figure~\ref{fig:fragments}.A shows miss rate vs. test set occurrences, identifying a few fragments that are disproportionately missed and labeling them with their respective molecular visualizations. Terminal chlorides are often missed; the plot specifically highlights aliphatic chlorides, but this is observed for ring substituents as well.  While this may be due in part to class imbalance, it is more likely the result of overlapping absorption ranges in the fingerprint region that are attributed to other components at the expense of these fragments. This is corroborated by the visualization in Figure~\ref{fig:fragments}.B, which shows the miss rate for select fragments for our model (IR only) and our model with formula conditioning. We can see that the inclusion of the formula in input eliminates much of the issue with the aforementioned terminal fragments, likely because the model is able to refer to the formula to distinguish overlapping spectra.

% ==============================================================
\section{Conclusion and Future Work}
\label{sec:conclusion}

In this work, we demonstrate various strategies for improving the performance of de novo molecular structure elucidation models. Our model achieves Top-1 accuracy of 31.83\%, Top-5 accuracy of 50.13\%, and Top-10 accuracy of 57.23\% on unseen experimental IR spectra. This is bolstered by the inclusion of fuzzy fusion of learned spectral features in the encoder and SMILES semantics in the decoder, as well as contrastive alignment. We find that fusion of different critical components in the transformer architecture is beneficial, suggesting that these models benefit from non-standard or fuzzy approaches to information aggregation in complex domains. We complement direct accuracy performance with additional metrics and visualizations that characterize robust capabilities of these models for analyzing infrared spectra and generating chemically valid SMILES strings. While accuracy compared to isomer ranking models is lower, we illustrate that the ultimate difference between generated candidates is minor, with the inclusion of formula priors only serving to distinguish underrepresented or overlapping absorption bands for small molecular components such as aryl thiols or terminal chlorides.

These findings ultimately serve to emphasize the degree to which IR spectra encode structural information, and that the application of machine learning has the potential to accelerate typical elucidation workflows while also making the most out of measured properties. Future work in this area could further explore fuzzy aggregation, potentially implementing new operators that better serve either the limited scope of available experimental data, or the specific chemical concepts learned by model sub-components such as decoder experts. Another avenue for future research involves the shared latent space that is formed through the contrastive loss term. We find high retrieval accuracy for test samples, further suggesting the existence of a chemically structured representation. This may benefit downstream tasks such as research into multi-modal models for structure elucidation that incorporate other spectral properties, or better characterization of graph-based approaches.

% ==============================================================
\section*{Resources}

We open source all Python code associated with the paper, including custom model implementations and training/evaluation scripts. The code will be available at \url{https://github.com/ethanmick8/ir2structure}. The experimental data are sourced from the NIST Chemistry WebBook (\url{https://webbook.nist.gov/chemistry/}). For the IBM-MD dataset \cite{alberts2024leveraging}, data can be obtained from \url{https://zenodo.org/records/7928396}, while the QM9S \cite{zou2023qm9s} was obtained from \url{https://figshare.com/articles/dataset/QM9S_dataset/24235333}.

% ==============================================================
\section*{Acknowledgments}

The authors acknowledge funding support from National Science Foundation (NSF) Award Number 2243526; NSF NRT-HDR: Advancing Materials Frontiers with Creativity and Data Science. M.J.Y. acknowledges partial support from the NSF Division of Materials Research (DMR) Polymers Program through Award Number 2235161. Additional support provided by the MU Materials Science \& Engineering Institute (MUMSEI). Computational resources were provided by the National Research Platform (NRP) Nautilus cluster \cite{weitzel2025nautilus}.
 
% ==============================================================
\bibliographystyle{unsrtnat}
\bibliography{references}

% ==============================================================
\appendix
 
% ==============================================================
\section{Implementation Details}
\label{supp:implementation}

%% ─────────────────────────────────────────────────────────────
\subsection{Architecture}
\label{supp:model_architecture}

Transformers are primarily based on the self-attention mechanism, which involves mapping a query ($Q$) and a set of key-value pairs ($K$,$V$) to an output. We calculate a scaled dot-product between $Q$ and $K$:\begin{equation} \begin{aligned} \text{Attention}(Q, K, V) = \text{softmax}(\frac{QK^{T}}{\sqrt{d_{k}}})V, \end{aligned} \label{eq:attention} \end{equation} where softmax is a normalization layer yielding a probability distribution over respective rows in the matrix. The result of self-attention layers is a global receptive field over data representation, a key advantage over previous recurrent modeling networks.

Multi-head attention extends the single-head attention mechanism by computing $h$ attention functions in parallel over learned linear projections of $Q$, $K$, and $V$: \begin{equation} \text{MultiHead}(Q, K, V) = \text{Concat}(\text{head}_1, \ldots, \text{head}_h)W^O, \label{eq:mha} \end{equation} where each head is computed as $\text{head}_{i}=\text{Attention}(QW_{i}^{Q},KW_{i}^{K},VW_{i}^{V})$, with learned projection matrices $W_{i}^{Q},W_{i}^{K} \in \R^{d_{model} \times d_k}, W_{i}^{Q} \in \R^{d_{model} \times d_v}$ and output projection $W^{O} \in \R^{hd_{v} \times d_{model}}$. The dimension per head is set to $d_{k} = d_{v} = d_{model}/h$, keeping total parameter cost constant relative to single-head attention at the same model width. Each head attends independently over the full sequence, allowing the model to jointly represent information from different representation subspaces at different positions. This is a property particularly relevant to IR spectra, where structurally informative features are distributed across disjoint spectral regions.

Each attention sub-layer is followed by a position-wise feed-forward network (FFN) applied identically and independently to each sequence position: \begin{equation} \text{FFN}(x) = \max(0, xW_1 + b_1)W_2 + b_2, \label{eq:ffn} \end{equation} where $W_1 \in \R^{d_{model} \times d_{ff}}, W_2 \in \R^{d_{ff} \times d_{model}}$, and $d_{ff} > d_{model}$ is the inner dimensionality, typically set to $4 \times d_{model}$. While the attention sub-layer captures dependencies across positions, the FFN applies a learned nonlinear transformation locally at each position, providing the representational capacity to encode complex token-level features. The encoder and decoder both follow this attention-then-FFN pattern per layer, with residual connections and layer normalization applied around each sub-layer. The replacement of this FFN with a Mixture of Experts (MoE) module is the modification described in Section \ref{sec:fuzzy_decoder}.

Since the attention mechanism is permutation-equivariant — it contains no inherent notion of token order — positional information must be injected explicitly. Transformer models typically use sinusoidal or learned absolute positional embeddings \cite{vaswani2017attention,radford2018improving}, which are appended to input embeddings prior to attention. In this work, we instead apply Rotary Position Embeddings (RoPE) \cite{su2023rope}, which encode position by rotating the query and key vectors in each attention head by an angle proportional to their absolute position prior to computing the scaled dot-product. This results in an augmented version of Equation~\ref{eq:attention}:
\begin{equation} 
\text{Attention}(Q, K, V) = 
\operatorname{softmax}\!\left(\frac{(R_\Theta^m Q)(R_\Theta^n K)^\top}{\sqrt{d_k}}\right)V, 
\label{eq:rope_attention} 
\end{equation} 
where $R_{\Theta}^{m}$ is a block-diagonal rotation matrix parameterized by position $m$ and a set of fixed frequency bases $\Theta$. The key property of this formulation is that the inner product $\langle R_{\Theta}^{m}q, R_{\Theta}^{n}k \rangle$ depends only on the relative offset $m-n$, meaning relative positional relationships are encoded implicitly through the rotation rather than added as a separate absolute embedding term. Compared to sinusoidal and learned absolute schemes, RoPE has been shown to extrapolate better to sequence lengths not seen during training and to integrate more naturally with the attention computation, since positional information is embedded directly into the query-key interaction rather than into the token representations themselves \cite{su2023rope}. It has also been shown to be effective for chemical language modeling and foundation models \cite{ross2022molformer}. RoPE is used in the self-attention modules of the architecture, but not the cross-attention, since in that case $Q$ comes from the decoder sequence while $K$ and $V$ are from the encoder, making position ambiguous.

%% ─────────────────────────────────────────────────────────────
\subsection{Fuzzy Aggregation Operators}
\label{supp:fuzzy_agg}

\paragraph{Linear Order Statistic Neuron (LOSN).} The LOSN \cite{veal2020losn} generalizes the Ordered Weighted Average (OWA) \cite{yager1988owa} to a neural setting by replacing the perceptron's index-aligned dot product with a weight vector applied to the \emph{sorted} inputs. For a scalar input vector $\vx_{\pi}$ sorted in descending order $x_{\pi(1)} \geq x_{\pi(2)} \geq \cdots \geq x_{\pi(N)}$, the LOSN is: 
\begin{equation} 
    f_{\text{LOSN}}(\mathbf{x}, \mathbf{w}) = 
    \mathbf{w}^t \mathbf{x}_\pi + w_{N+1},
    \label{eq:losn} 
\end{equation} 
where $\mathbf{w} \in \R^{N+1}$ are learned weights and $w_{N+1}$ is a bias. The LOSN can be interpreted as $N!$ shared weight perceptrons — one per possible input ordering — collapsing the exponential number of sort-order configurations into a single parameter vector through the shared weights. Critically, by learning $\mathbf{w}$ rather than fixing it, the LOSN can recover max($w_1 = 1, w_{k>1} = 0$), min($w_N = 1, w_{k<N} = 0$), mean($w_k = 1/N$), and continuously interpolated variants thereof, providing a smooth, learnable spectrum between extremal and central aggregation behaviors.

The Extended LOSN \cite{kakula2020elos} introduces an additional layer of expressivity by applying LOSN independently at each output dimension rather than with a single shared weight vector across all dimensions. This permits a model configuration to learn to distinguish different dimensions of input representations in addition to their primary differences.

\paragraph{Choquet Integral Multilayer Perceptron (ChIMP).} The discrete Choquet integral of outputs $\mathbf{h} = (h_1,...,h_{E})$ with respect to $g$ is: 
\begin{equation} 
    C_g(\mathbf{h}) = 
    \sum_{j=1}^{E} h_{\pi(j)}\left(g(A_{\pi(j)}) - g(A_{\pi(j-1)})\right),
    \label{eq:choquet} 
\end{equation} 
for permutation $\pi$ such that $h_{\pi(1)} \geq h_{\pi(2)} \geq \cdots \geq h_{\pi(E)}$, and where $A_{\pi(j)} = x_{\pi(1)},...,x_{\pi(j)}$ and $A_{\pi(j)} = 0$. The FM $g$ encodes the worth of each data source coalition: $g(\{e\})$ reflects the marginal value of source $e$ alone, while $g(\{e_1,e_2\})$ captures synergistic or redundant interactions between pairs, with the Interaction Index $\mathcal{I}_{g}(i,j) \in [-1,1]$ quantifying whether combining two sources is complementary (+1) or redundant (--1). Islam \textit{et al.} \cite{islam2020chimp} prove that the Choquet Integral (ChI) admits a multilayer network representation (ChIMP) that can be optimized via stochastic gradient descent (SGD) using an iChIMP formulation based on the M{\"o}bius representation of the integral: 
\begin{equation} 
    C_g(\mathbf{h}) = \sum_{A \subseteq X} g(A)o(A), 
    \quad o(A) = 
    \begin{cases} 
        \max\left(0, \bigwedge_{x_i \in A} h_i - \bigvee_{x_j \notin A} h_j\right) & A \subset X,\\
        \bigwedge_{x_i \in A}{h_i} & A = X.,
    \end{cases} 
    \label{eq:ichimp} 
\end{equation} 
where the integrand terms $o(A)$ are computed by a fixed sub-network of max, min, and $f(a,b) = \text{max}(0,a-b)$ neurons with no learnable weights, and the FM variables $g(A)$ are the sole learned parameters. The full iChIMP formulation maintains $2^{E}-1$ variables and computes $2^{E}-1$ integrand terms, which becomes intractable as $E$ grows. In this work we adopt the $k=2$ additive restriction \cite{grabisch1997kadditive}, which approximates the full FM by retaining only singleton and pairwise subset terms while setting all higher-order coalition values to zero. Under this restriction the Choquet integral reduces to:
\begin{equation}
  C_g^{(2)}(\mathbf{h}) = \sum_{e=1}^{E} \alpha_e \cdot h_{\pi(e)} + \sum_{i < j} \beta_{ij} \cdot \min\left(h_{\pi(i)}, h_{\pi(j)}\right),
  \label{eq:chimp_2additive}
\end{equation}
where $\alpha_e \in \mathbb{R}$ are learned singleton weights, $\beta_{ij} \in \mathbb{R}$ are learned pairwise interaction weights, and the $\min$ operator is the standard $t$-norm realization of the pairwise interaction term under 2-additivity. The parameter count reduces from $2^{E}-1$ to $E + \binom{E}{2}$ while retaining the full pairwise interaction structure that is the primary expressiveness advantage of ChIMP over linear aggregation. This eliminates the need for the iChIMP fixed sub-network for computed $o(A)$ terms; the 2-additive integral is computed directly as the sum of singleton-weighted sort inputs and interaction-weighted element-wise minima of sorted input pairs. Singleton weights are initialized at $1/E$ and interaction weights at zero, recovering a uniform weighted mean at initialization and allowing pairwise structure to emerge only as training supports it. An $\ell_1$ regularization term on $\beta$ encourages sparse interaction structure, retaining only robustly supported source-pair relationships \cite{kakula2020choquetreg}.

\paragraph{Load balancing for routed aggregation.} The order-statistic and Choquet operators above aggregate over the full set of expert outputs independently of the router. The sparse top-$k$ LCS baseline instead aggregates only the experts the router selects, which admits a failure mode absent from the fuzzy variants: without intervention the router can collapse onto a small subset of experts, routing nearly all tokens to them and leaving the rest undertrained. Following Fedus \textit{et al.} \cite{fedus2022switchtransformer}, the baseline is regularized with an auxiliary load-balancing loss
\begin{equation}
  \mathcal{L}_{\text{bal}} = E \sum_{e=1}^{E} f_e \cdot P_e,
  \label{eq:load_balance}
\end{equation}
where $E$ is the number of experts, $f_e$ is the fraction of batch tokens dispatched to expert $e$, and $P_e$ is the mean router probability assigned to expert $e$ over the batch. The product $f_e P_e$ is minimized when both dispatch and routing confidence are spread uniformly, so the term penalizes imbalance; it is added to the objective with a small coefficient, $\lambda_{bal}$, and disabled at inference. Because MoE-LOSN and MoE-ChIMP aggregate over all expert outputs rather than a router-selected subset, no expert is starved of gradient, and these variants are trained without this auxiliary term.

%% ─────────────────────────────────────────────────────────────
\subsection{Model Hyperparameters}
\label{supp:model_hparams}

This section details the specific architecture hyperparameter settings employed for trained models. All experiments use the configuration settings detailed in Table~\ref{tab:supp:encoder} and Table~\ref{tab:supp:decoder} unless otherwise ablated. We keep the number of free/trainable parameters for the model fixed around 33 million; any variation here is due to the negligible addition of small layers for different architectural variations, such as the weights included in the LOSN encoder layer fusion as described in Equation~\ref{eq:losn_fusion}.

\begin{table}[h!]
\centering
\caption{Spectral encoder hyperparameters.}
\label{tab:supp:encoder}
\begin{tabular}{lll}
\toprule
\textbf{Hyperparameter} & \textbf{Value} & \textbf{Description} \\
\midrule
\multicolumn{3}{l}{\textit{Spectral pre-processing}} \\
Wavenumber range        & 450--3{,}981 cm$^{-1}$   & Input spectral domain \\
Spectral resolution     & 1{,}791 points           & Points after resampling \\
Patch size              & 75                       & Consecutive points per patch \\
Number of patches       & 23                       & Sequence length into encoder \\
\midrule
\multicolumn{3}{l}{\textit{Transformer encoder}} \\
$d_\text{model}$        & 512                      & Model / embedding dimension \\
num\_heads              & 8                        & Attention heads \\
$d_\text{head}$         & 64                       & Dimension per head \\
$N_\text{enc}$          & 3                        & Number of encoder layers \\
$d_\text{ffn}$          & 2{,}048                  & FFN hidden dimension \\
Positional encoding     & RoPE                     & Rotary position embedding \cite{su2023rope} \\
Normalization           & Post-norm                & LayerNorm after residual \cite{ba2016layernormalization} \\
Dropout                 & 0.05                     & Applied to attention and FFN \\
\midrule
\multicolumn{3}{l}{\textit{Fuzzy Encoder Fusion}} \\
Fusion operator         & 2-additive Choquet       & Over $N_\text{enc}$ layer outputs \\
Include embedding layer & \texttt{False}           & Layers only, no patch embed \\
Gate input              & Layer mean pool          & $\mathbb{R}^{d_\text{model}}$ \\
Singleton weights init  & Uniform $1/L$            & \\
Pairwise weights init   & 0                        & Additive measure at init \\
Number of interaction terms & $\binom{N_\text{enc}}{2} = 15$ & \\
\midrule
\multicolumn{3}{l}{\textit{Contrastive Loss Terms}} \\
Loss                        & InfoNCE                  & \cite{oord2019infonce}; Symmetric spectral--structural \\
$\Loss_{align}$ weight      & $\lambda=1.0$            & Compound loss weighting \\
Temperature $\tau$          & 0.07                     & Learnable, initialized at 0.07 \\
Spectral pool method        & Attention pool           & per CoCa \cite{yu2022coca} \\
Projection dimension        & 512                      & Shared embedding space \\
Negatives                   & In-batch                 & \\
\bottomrule
\end{tabular}
\end{table}

\begin{table}[h!]
\centering
\caption{SMILES decoder hyperparameters.}
\label{tab:supp:decoder}
\begin{tabular}{lll}
\toprule
\textbf{Hyperparameter} & \textbf{Value} & \textbf{Description} \\
\midrule
\multicolumn{3}{l}{\textit{Tokenization}} \\
Tokenizer               & Atom-level SMILES        & Character + atom-type tokens \cite{schwaller2019regex} \\
Vocabulary size         & 47                       & Comprising all possible tokens in explored data\\
Max SMILES length       & 100                      & Tokens including \texttt{[BOS]}/\texttt{[EOS]} \\
\midrule
\multicolumn{3}{l}{\textit{Transformer decoder}} \\
$d_\text{model}$        & 512                      & Shared with encoder \\
num\_heads              & 8                        & Attention heads \\
$d_\text{head}$         & 64                       & Dimension per head \\
$N_\text{dec}$          & 4                        & Number of decoder layers \\
$d_\text{ffn}$          & 2{,}048                  & FFN hidden dimension \\
Positional encoding     & RoPE                     & \\
Self-attention          & Causal (masked)          & \\
Cross-attention         & Standard MHA             & Query: decoder; Key/Value: encoder \\
Normalization           & Post-norm                & \\
Dropout                 & 0.1                      & \\
\midrule
\multicolumn{3}{l}{\textit{Decoder MoE (when active)}} \\
\texttt{num\_experts}   & 8                                     & \\
\texttt{top\_k}         & 4                                     & \\
Expert sizes            & 512                                   & Matching active parameters of dense model \\
Aggregation             & LCS / LOSN / ChIMP                    & Per ablation condition \\
$\Loss_{bal}$           & $\lambda_{bal}=0.01$                  & Load balancing loss weight (MoE-LCS only) \\
ChIMP group size        & 64                                    & Features per Choquet group \\
Number of groups        & $d_\text{model}/64 = 8$               & ChIMP specific \\
\midrule
\multicolumn{3}{l}{\textit{Inference}} \\
Decoding                & Beam search              & \\
Beam width              & 10                       & Number of generated candidates \\
Length penalty          & 1.0                      & \\
\bottomrule
\end{tabular}
\end{table}

%% ─────────────────────────────────────────────────────────────
\subsection{Training Settings}
\label{supp:training}

We train models on a single NVIDIA A100-80GB GPU (PCIe). The CPU used is the AMD EPYC 7713 (64 cores, 2.00 GHz, 256 cache), with 150 GB of RAM.

We conduct a standard ablation of training hyperparameters to identify the optimal configuration for trained models. We utilize ADAM optimization, a 1-cycle learning rate schedule with a learning rate of 5e-4 and a 30\% warmup phase, as well as a batch size of 256 for pretraining and 64 for finetuning.

% ==============================================================
\section{Molecular Fragment Analysis}
\label{supp:fragments}

%% ─────────────────────────────────────────────────────────────
\subsection{Custom Fragment Decomposition}
\label{supp:fragments_custom}

To evaluate structural recovery beyond exact-match and fingerprint-based metrics, we decompose both ground-truth and predicted SMILES into sub-structural fragments using a custom aliphatic-linker fragmentation engine.

\paragraph{Fragmentation algorithm.}
Each molecule is fragmented by first identifying sp\textsuperscript{3} aliphatic carbons (non-ring, single-bond only) as cut points, followed by a breadth-first search (BFS) from non-linked neighbors to yield distinct regions. A second pass is then conducted to strip substituents from ring cores and produce isolated ring scaffolds and their R-groups. The result is five semantically meaningful fragment classes: \textit{Aromatic}, consisting of intact aromatic ring systems (benzene, pyridine), \textit{Non-Aromatic Ring}, consisting of saturated/partially unsaturated rings such as cyclohexane, \textit{Aromatic Substituent}, consisting of R-groups directly attached to aromatic rings, \textit{Non-Aromatic Ring Substituent}, same as the previous but for non-aromatics, and \textit{Aliphatic}, which collects free sp\textsuperscript{3} chains/branches not adjacent to ring systems. Fragment identity is therefore represented as the tuple $(\text{canonical SMILES},\, \text{label})$; two fragments are considered equivalent only when both components agree.

\paragraph{Per-sample metrics.}
Let $\mathcal{G}$ and $\mathcal{P}$ denote the fragment sets of the ground-truth and predicted molecules, respectively.  We define

\begin{equation}
  \text{Recall} = \frac{|\mathcal{G} \cap
\mathcal{P}|}{|\mathcal{G}|},
  \qquad
  \text{Precision} = \frac{|\mathcal{G} \cap
\mathcal{P}|}{|\mathcal{P}|},
  \qquad
  F_1 = \frac{2 \cdot \text{Recall} \cdot \text{Precision}} {\text{Recall} + \text{Precision}},
\end{equation}

with the convention that recall $= 1$ when $|\mathcal{G}| = 0$, and precision $= 1$ (resp.\ $0$) when $|\mathcal{P}| = 0$ and $|\mathcal{G}| = 0$ (resp.\ $|\mathcal{G}| > 0$).  For models that generate $K$ beam hypotheses per spectrum, we select the beam that maximises $(\text{Recall},\, F_1)$ lexicographically.  Macro-averaged recall, precision, and $F_1$ are then reported across the test set, alongside a \textit{Molecular Perfection Rate} (MPR) — the fraction of samples for
which the selected beam contains neither missing nor extraneous fragments relative to the ground truth.

\paragraph{Fragment difficulty analysis.}
To identify structurally challenging motifs, we accumulate per-fragment shared and missed counts across the test set and restrict attention to fragments appearing in at least $N_{\min} = 10$ ground-truth molecules and above the 50\textsuperscript{th} percentile of occurrence frequency. Each qualifying fragment is ranked by miss rate, enabling systematic identification of motifs that the model consistently fails to recover. This is the primary process that is applied and visualized in Figure~\ref{fig:fragments}.

%% ─────────────────────────────────────────────────────────────
\subsection{BRICS Fragment Analysis}
\label{supp:fragments_brics}

We also decompose molecules using the BRICS algorithm of Degen \textit{et al.} \cite{degen2008brics}, which defines 16 retrosynthetically motivated bond-environment rules derived from medicinal-chemistry practice.  Fragments are identified solely by their canonical SMILES string (BRICS dummy-atom labels $[\mathrm{n*}]$ are retained during canonicalization but stripped before structural classification); per-sample recall, precision, and $F_1$ are computed identically to the custom scheme above.

\paragraph{Structural category classification.}
To obtain chemically interpretable error breakdowns, each BRICS fragment is assigned to one or more structural categories via SMARTS substructure matching (Table~\ref{tab:brics_categories}).  Classification is \textit{multi-label}: a fragment may belong to several categories simultaneously (i.e., a pyridine ring is both \textit{aromatic ring} and \textit{nitrogen heterocycle}). Category-level true positives (TP), false positives (FP), and false negatives (FN) are accumulated over all test samples and used to compute per-category precision, recall, and $F_1$, as reported in Figure~\ref{fig:brics}.

\begin{table}[h]
\centering
\small
\caption{Structural categories used for BRICS fragment classification. Matching is applied after stripping numbered dummy labels
($[\mathrm{n*}] \!\to\! [*]$).  Atom number \texttt{\#0} denotes a dummy attachment atom and is used in the ether/ester pattern to capture fragments in which one or both oxygen neighbours are attachment points rather than explicit carbons.}
\label{tab:brics_categories}
\begin{tabular}{lll}
\toprule
\textbf{Category} & \textbf{Detection criterion} & \textbf{Notes} \\
\midrule
Aromatic ring
  & Any non-dummy heavy atom is aromatic
  & \\
Nitrogen heterocycle
  & \texttt{[n,N;R]}
  & Aromatic or aliphatic N in a ring \\
Amide / lactam
  & \texttt{[CX3](=O)[NX3,NX2]}
  & Supersedes carbonyl (non-amide) \\
Carbonyl (non-amide)
  & C\texttt{=}O double bond; no amide match
  & Aldehydes, ketones, carbamates \\
Ether / ester linkage
  & \texttt{[\#6,\#0][OX2;!H][\#6,\#0]}
  & Bridging non-hydroxylic O \\
Hydroxyl / alcohol
  & \texttt{[OX2H]}
  & Free O–H \\
Halogen-bearing
  & \texttt{[F,Cl,Br,I]}
  & Any halogen substituent \\
Aliphatic chain ($\geq$3\,C)
  & $\geq$3 non-aromatic, non-ring sp$^3$ carbons
  & All bonds single
\\
Other
  & No category matched
  & Catch-all \\
\bottomrule
\end{tabular}
\end{table}

% ==============================================================
\section{Additional Results}
\label{supp:addl_results}

%% ─────────────────────────────────────────────────────────────
\subsection{Mixture of Experts and Token Attribution}
\label{supp:moe}

\begin{table}[h]
\centering\small
\caption{SMILES token categories used for expert attribution. Each character position of the SMILES string is assigned exactly one category. Atom ring membership and aromaticity are determined by RDKit \cite{rdkit} (\texttt{GetIsAromatic}, \texttt{RingInfo}), with a string-only heuristic fallback for SMILES that fail to parse; multi-character atoms (e.g., \texttt{Cl}, bracketed atoms) assign the atom's category to every character in its span. Control tokens (\texttt{[BOS]}/\texttt{[EOS]}/padding) are excluded prior to classification via the decoder token offset.}
\label{tab:token_categories}
\begin{tabular}{ll}
\toprule
\textbf{Category} & \textbf{Detection criterion} \\
\midrule
\multicolumn{2}{l}{\textit{Structural / connectivity markers}} \\
\texttt{branch\_open}   & \texttt{(} \\
\texttt{branch\_close}  & \texttt{)} \\
\texttt{ring\_closure}  & Ring-bond digit or \texttt{\%dd} \\
\texttt{double\_bond}   & \texttt{=} \\
\texttt{triple\_bond}   & \texttt{\#} \\
\texttt{aromatic\_bond} & \texttt{:} \\
\texttt{stereo}         & \texttt{@}, \texttt{@@}, \texttt{/}, \texttt{\textbackslash} \\
\midrule
\multicolumn{2}{l}{\textit{Carbon}} \\
\texttt{C\_aromatic}    & Aromatic carbon \\
\texttt{C\_ring}        & Aliphatic carbon in a ring \\
\texttt{C\_chain}       & Aliphatic carbon not in a ring \\
\midrule
\multicolumn{2}{l}{\textit{Nitrogen}} \\
\texttt{N\_aromatic}    & Aromatic nitrogen \\
\texttt{N\_ring}        & Aliphatic nitrogen in a ring \\
\texttt{N\_chain}       & Aliphatic nitrogen not in a ring \\
\midrule
\multicolumn{2}{l}{\textit{Oxygen / Sulfur}} \\
\texttt{O\_ring}        & Oxygen in a ring \\
\texttt{O\_chain}       & Oxygen not in a ring \\
\texttt{S\_ring}        & Sulfur in a ring \\
\texttt{S\_chain}       & Sulfur not in a ring \\
\midrule
\multicolumn{2}{l}{\textit{Halogen}} \\
\texttt{F}              & Fluorine \\
\texttt{Cl}             & Chlorine \\
\texttt{Br}             & Bromine \\
\texttt{I}              & Iodine \\
\midrule
\multicolumn{2}{l}{\textit{Other}} \\
\texttt{P}              & Phosphorus \\
\texttt{other\_atom}    & Any other atomic symbol \\
\texttt{other}          & Any non-atom, non-structural character \\
\bottomrule
\end{tabular}
\end{table}

\begin{figure}[h!]
    \centering
    \includegraphics[width=1.0\textwidth]{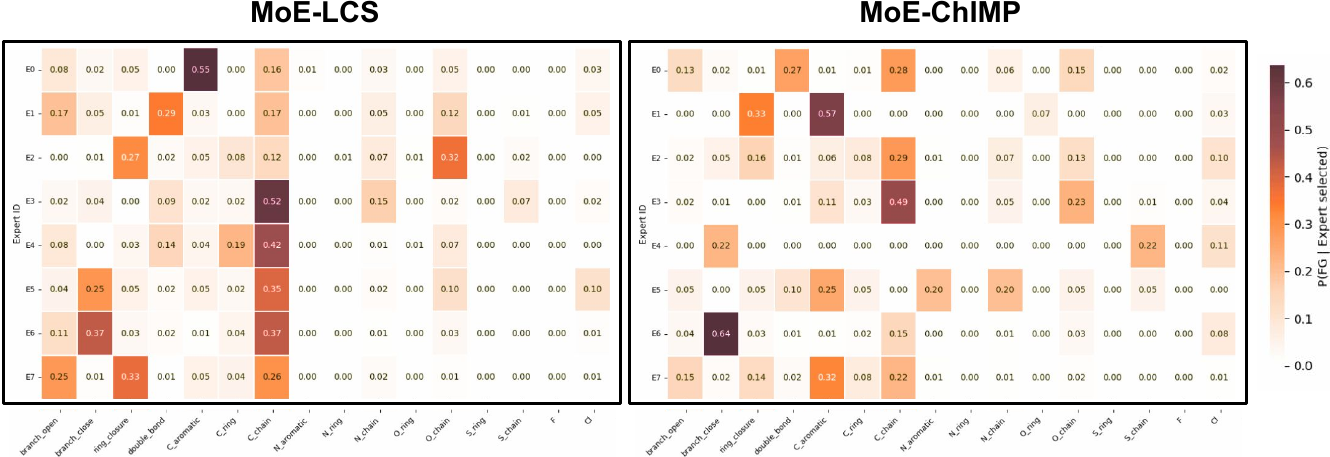}
    \caption{Expert SMILES token attribution for different MoE configurations. Bin intensity denotes the probability that a token belongs to a SMILES group given that the corresponding expert (row) was selected for it. C\_chain is the most common of these groups; MoE-LCS makes this clear by delegating it to several experts. MoE-ChIMP, on the other hand, aided by the contribution of its informed aggregation module, disperses focus more evenly across potential classes.}
    \label{fig:expert_token_attrib}
\end{figure}

Figure~\ref{fig:expert_token_attrib} reports, for each expert, the distribution of SMILES token categories over the positions where that expert was the gate's top choice. At every decoder MoE position, the gate produces a confidence ranking over the $E$ experts; we record the rank-0 expert (highest gate confidence). This definition is applied identically to all variants. Each position is mapped back to its SMILES character and assigned to one of the [N] token categories of Table~\ref{tab:token_categories}; positions corresponding to control tokens (\texttt{[BOS]}/\texttt{[EOS]}/padding) are excluded. Counts are accumulated over the test set into an $E \times [N]$ matrix and row-normalized, so entry $(e,c)$ gives $P(\text{category } c \mid \text{expert } e \text{ is rank-0})$ and each expert row sums to one.

We attribute on the gate's top choice rather than the realized aggregation weight so that the metric is well-defined and comparable across variants. For the routed LCS baseline, the rank-0 expert also participates directly in the convex sum; for MoE-LOSN and MoE-ChIMP the gate is computed but does not drive aggregation, which instead sorts and combines all expert outputs (Section~\ref{sec:fuzzy_decoder}). The heatmap therefore characterizes the specialization the \emph{gate} has learned under each aggregation objective, not the per-token contribution to the output. 

We see that non-additive aggregation (MoE-ChIMP) appears to lead to more diverse experts, potentially placing a greater emphasis on under-represented substructures. The MoE-LCS, on the contrary, sees several of its experts focused on carbon chains (C\_chain), which is easily the most common of the measured token types as defined in Table~\ref{tab:token_categories}. This dispersion may follow directly from the gate decoupling; and for ChIMP the unconstrained organization may be further amplified: because the 2-additive measure assigns worth to \emph{pairs} of experts, an individual expert need not be a self-sufficient predictor of any one category but can specialize narrowly and contribute through its interaction terms with complementary experts, rewarding coverage of under-represented motifs that are informative only in combination.

%% ─────────────────────────────────────────────────────────────
\subsection{Encoder Fusion}
\label{supp:encoder_fusion}

\begin{figure}[h!]
    \centering
    \includegraphics[width=1.0\textwidth]{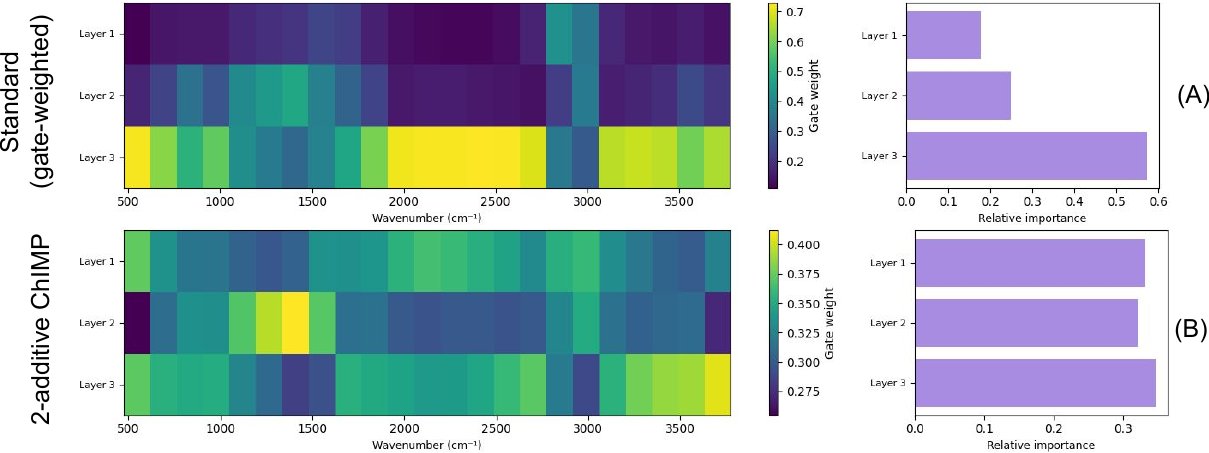}
    \caption{Encoder Fusion layer weight by spectral position, and relative contribution, for the standard gate-weighted approach and the proposed 2-additive ChIMP method. Higher gate weight indicates that the corresponding layer dominates at that wavenumber.}
    \label{fig:supp:encoder_fusion}
\end{figure}

To further characterize the effect that fuzzy fusion operators have on encoder layer utilization, Figure~\ref{fig:supp:encoder_fusion} presents a comparative analysis of the standard gate-weighted sum and the 2-additive Choquet integral fusion modes applied to the stacked encoder layer representations produced at each depth level of the spectral encoder. For each fusion variant we visualize (left panels) a heatmap of normalized per-layer contribution as a function of wavenumber $\tilde{\nu}$ (cm$^{-1}$), where each row corresponds to one encoder depth and color intensity reflects the effective weight assigned to that layer at each spectral position, and (right panels) the marginal contribution averaged across $\tilde{\nu}$, summarizing aggregate layer utilization.

The 2-additive Choquet integral, by contrast, sorts encoder layer outputs at each spectral position by their L2-norm magnitude and applies learned rank-position weights rather than learned layer-identity weights. Because the identity of the highest-norm layer varies across $\tilde{\nu}$, the effective per-layer contribution becomes position-dependent without any spatial supervision. Figure~\ref{fig:supp:encoder_fusion}.B demonstrates the resulting behavior: contribution is balanced across all encoder depths, and each layer displays a distinct spectral focus region corresponding to the wavenumber range where its representations carry the highest relative magnitude. With the 2-additive choquet integral, each pairwise term activates most strongly at positions where two rank-consecutive layers are simultaneously high-norm, producing conjunctive bonus at spectrally rich positions where multiple depths of encoding contribute complementary evidence. Ultimately, we get a depth-resolved, spectrally localized decomposition of encoder information that the standard gate-weighted mechanism cannot express.

%% ─────────────────────────────────────────────────────────────
\subsection{SMILES Augmentation Analysis}
\label{supp:smiles_aug}

\begin{figure}[h!]
    \centering
    \includegraphics[width=1.0\textwidth]{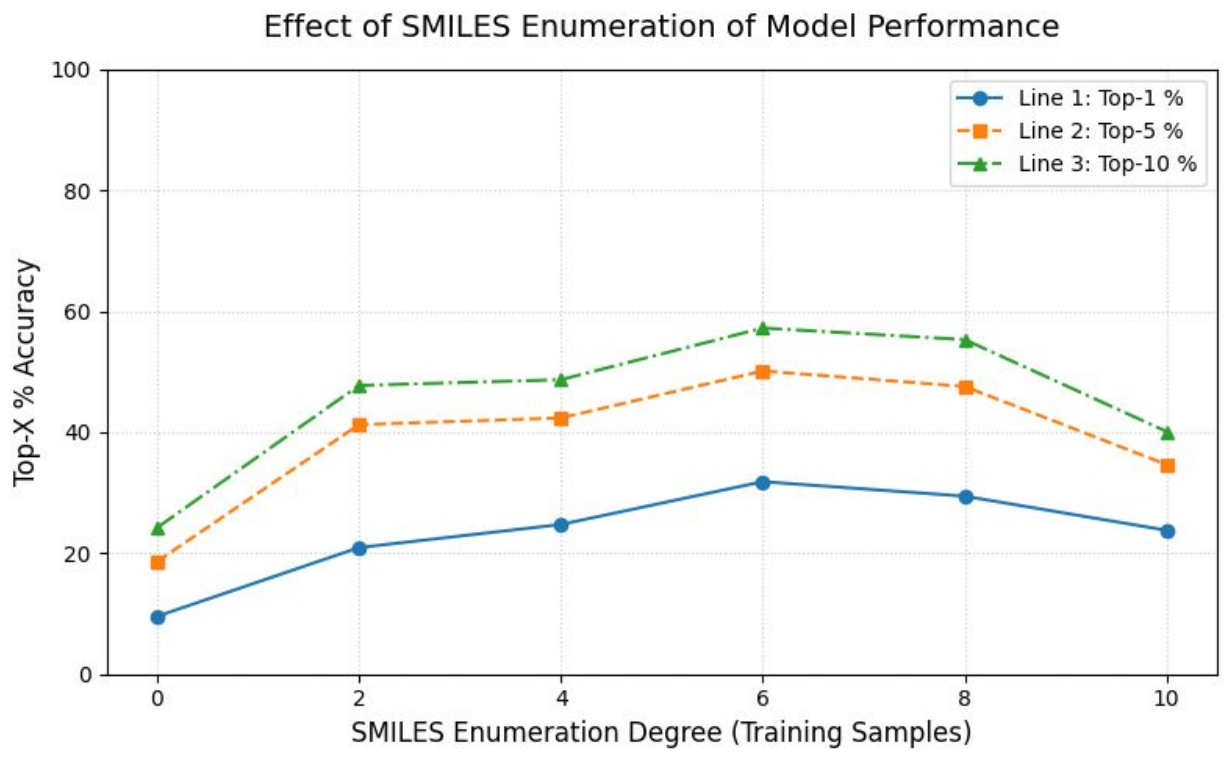}
    \caption{Varying degrees of SMILES enumeration of training samples on the experimental training set before fine-tuning, and their corresponding performance. Increasing degree of enumeration increases the size of the fine-tuning dataset.}
    \label{fig:supp:smiles_aug}
\end{figure}

Introduced in \cite{bjerrum2017smilesenum} and applied to isomer ranking models in Alberts \textit{et al.} \cite{alberts2025setting}, SMILES augmentation increases the size of the dataset by enumerating chemically equivalent SMILES representations for training samples. Previous works find that doubling the size of the training dataset by enumerating a single chemically equivalent SMILES string for each sample enhances Top-1 accuracy. We extend this analysis to explore the effect of additional enumeration, as most molecules have several chemically equivalent representations in addition to the canonical string. Figure \ref{fig:supp:smiles_aug} details the effect of different degrees of enumeration on the fine-tuning training corpus. The experimental test set remains unchanged. The model utilized is the best performing model as identified in the main body, save for the degree of augmentation.

The primary benefit of SMILES augmentation is forcing the decoder to treat the structural prediction problem as representation-invariant, and this is better represented when the decoder is given a diverse range of SMILES-space representations to attribute to spectra. In Figure~\ref{fig:supp:smiles_aug}, we see the effect of different degrees of SMILES augmentation on our model. The decoder's implicit prior over SMILES syntax is flatter and more robust with the added enumerations, and the impact is apparent — including no augmentations results in dramatically lower Top-K accuracy. As we add augmentations, this stabilizes. Ultimately, the specific enumeration value chosen is going to depend on the overall complexity of training molecules; for the relatively small experimental NIST set, this is determined to be 6. At 8 augmentations, we start to observe a dropoff in performance. This may be due to the contrastive loss, which specifically treats in-batch pairs as negatives. With each spectrum appearing 8 times in the training data, a non-trivial fraction of batches may contain multiple augmentations of the same molecule, leading the loss to push spectral embeddings \textit{away} from SMILES embeddings that are actually correct structural matches.
 
\end{document}